\documentclass[11pt]{article}

\usepackage[utf8]{inputenc}
\usepackage[a4paper, margin=1in]{geometry}
\usepackage{graphicx}
\usepackage{amsmath}
\usepackage{hyperref}
\usepackage[table]{xcolor}
\usepackage{wrapfig}
\usepackage{booktabs}
\usepackage{longtable}
\usepackage{xcolor}
\usepackage{authblk}
\usepackage{lineno}
\usepackage[singlespacing]{setspace}
\usepackage{comment}
\usepackage{float}
\usepackage{makecell}
\usepackage{booktabs} 
\usepackage{subcaption}
\usepackage{siunitx}
\usepackage{tabularx}
\usepackage{wrapfig}
\usepackage{placeins}
\newcolumntype{L}[1]{>{\raggedright\arraybackslash}p{#1}}
\newcolumntype{T}[1]{>{\ttfamily\raggedright\arraybackslash}p{#1}}
\renewcommand{\arraystretch}{1.5}
\usepackage{booktabs}
\usepackage{adjustbox}
\usepackage{multirow}
\usepackage{siunitx}
\usepackage{colortbl}
\usepackage{xcolor}

\usepackage{xcolor}

\title{\textbf{PlantExpertVQA: A Visual Question Answering Dataset for Benchmarking Vision-Language Models in Plant Science}}

\author[1]{Syed Nazmus Sakib}
\author[1]{Nafiul Haque}
\author[2]{Mohammad Zabed Hossain}
\author[1]{Shifat E. Arman}

\affil[1]{Department of Robotics and Mechatronics Engineering, University of Dhaka}
\affil[2]{Department of Botany, University of Dhaka}

\date{}
\begin{document}
\maketitle

\renewcommand{\thefootnote}{\fnsymbol{footnote}}  
\footnotetext[1]{Corresponding author: Shifat E. Arman (shifatearman@du.ac.bd)}
\footnotetext[2]{Co-corresponding author: Mohammad Zabed Hossain (zabed@du.ac.bd)}
\renewcommand{\thefootnote}{\arabic{footnote}}     
\renewcommand{\thefootnote}{}
\footnotetext{This article has been accepted for publication in \emph{Scientific Data} (Nature Portfolio).}
\renewcommand{\thefootnote}{\arabic{footnote}}

\begin{abstract}
Existing plant-disease datasets target classification and detection, leaving vision–language models unable to support interactive, reasoning-based diagnosis. To address this, we present \textbf{PlantExpertVQA}, a large-scale visual question answering (VQA) dataset designed to advance vision–language models for agricultural decision-making. It is compiled from 45 open-source datasets, including the widely used PlantVillage corpus, and comprises 765,186 high-quality question–answer (QA) pairs grounded over 150,841 images spanning 38 crop species and 89 disease conditions. Questions are organized into 3 levels of cognitive complexity and 9 distinct categories. Each was phrased following expert guidance and generated via an automated two-stage pipeline: template-based QA synthesis from image metadata, followed by multi-stage linguistic re-engineering. The dataset was iteratively reviewed by domain experts for scientific accuracy and relevance. We find that current frontier vision–language models, including recent open-source instruction-tuned multimodal LLMs, perform poorly on PlantExpertVQA. However, parameter-efficient fine-tuning of a compact 2B-parameter model on a small fraction of the dataset yields substantial improvements across all question categories, demonstrating its effectiveness for domain adaptation.
\end{abstract}

\section*{Background \& Summary}

Plant diseases threaten global food security and farm productivity. Research indicates that plant pests and diseases are responsible for the loss of as much as 30\% of global food crop yields annually \cite{savary2019global}. This results in famine, malnutrition, and food insecurity for hundreds of millions of people worldwide. In most cases pest and fungal invasions spread rapidly due to diagnostic delay. Therefore, precision tools are now needed for early symptoms detection and targeted interventions.

Machine learning has advanced plant pathology by enabling automated identification of disease symptoms. Convolutional neural networks and transformer models can reliably detect leaf diseases in various crop species \cite{mohanty2016using, ferentinos2018deep, bhuiyan2023bananasqueezenet,hossain2024deep}. However, most existing frameworks focus only on classification and do not provide insights into symptom causation or context.

Visual Question Answering (VQA) \cite{antol2015vqa, yang2016stacked} combines image understanding with natural language processing to answer queries about visual content. VQA databases go beyond classification by allowing interactive question–answering \cite{changpinyo2021conceptual}. This allows trained models to capture complex relationships in the images. As such, the application of VQA now spans multiple domains. These include: educational tools \cite{banerjee2022vqa, huang2022aitutors}, customer service systems \cite{jain2021customersupport, zhu2023ecommerce}, and autonomous driving \cite{marcu2024lingoqa} etc. In particular, VQA shows exceptional potential in the field of pathological diagnosis and health inquiry \cite{yan2023radvqa, zhang2023pmc}. Current medical VQA benchmarks include PMC-VQA \cite{zhang2023pmc}, SLAKE \cite{liu2021slake}, Path-VQA \cite{he2020pathvqa}, and VQA-RAD \cite{lau2018dataset}. However, these datasets are focused on medical diagnostics.

\begin{figure}[H]
    \centering
    \includegraphics[width=0.95\linewidth]{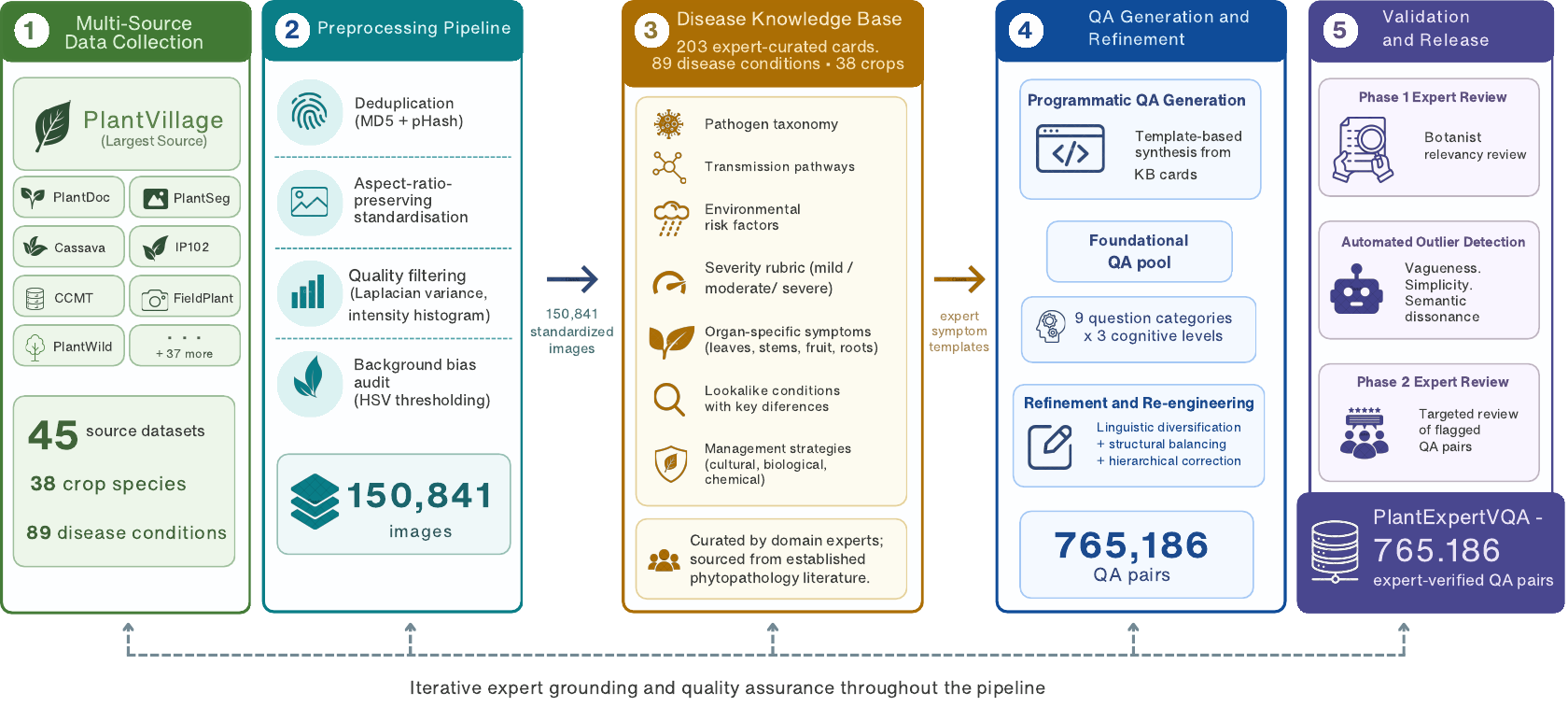}
    \caption{Overall Methodology of PlantExpertVQA creation}
    \label{fig: overall process}
\end{figure}

In agriculture, existing popular datasets like PlantVillage \cite{mohanty2016using}, PlantDoc \cite{singh2020plantdoc}, and PlantSeg \cite{wolny2020accurate} focus on classification or segmentation tasks. While they support disease detection, they do not enable interactive reasoning through question–answer formats. Recent systems like AgroGPT \cite{awais2025agrogpt}, LLaVa-PlantDiag \cite{sharma2024llava} incorporate VQA models with the PlantVillage dataset. But these resources use generalized large language models for text generation and lack rigorous expert verification.

To address these gaps, we introduce PlantExpertVQA, a domain-specific visual question answering dataset for plant disease diagnosis. The dataset is built on a compilation of 45 open-sourced datasets featuring explicit, well-curated hierarchical annotations that underwent extensive preprocessing and standardization. It contains 765,186 question–answer pairs grounded over 150,841 images across 38 unique crop species and 89 disease conditions, covering nine distinct question categories under three levels of cognitive complexity. Each question was naturally phrased and tailored through expert review. Our methodology combines automated template-based QA generation with multistage linguistic re-engineering and iterative botanist review to ensure clinical accuracy and domain relevance. We further benchmark the dataset by evaluating nine open-source vision–language models in a zero-shot setting, and demonstrate the dataset's utility as a training resource through parameter-efficient fine-tuning of a compact multimodal model. To our knowledge PlantExpertVQA is the first multimodal dataset created under extensive guidance by domain experts for the sole purpose of model training and evaluation. 

Overall, our primary contributions include:
\begin{enumerate}
    \item \textbf{Large-scale, expert-verified VQA dataset grounded in a structured Disease Knowledge Base}: We introduce PlantExpertVQA, a multi-source domain-specific VQA dataset consisting of 765,186 expert-verified QA pairs grounded over 150,841 images, covering 38 crop species and 89 disease conditions, sourced from 45 open-access repositories under verified permissive licenses. The dataset is built on a structured Disease Knowledge Base, in which every crop–condition pair is encoded as a machine-readable card capturing pathogen taxonomy, transmission pathways, environmental risk factors, severity rubrics, organ-specific visual symptoms, visually similar lookalike conditions with key differentiating features, and management strategies. The dataset spans nine question categories organized into three levels of cognitive complexity, ranging from basic identification to causal and counterfactual reasoning.
    
    \item \textbf{Two-Stage QA Generation Pipeline with Domain Expert Review}: We implement an automated template-based QA synthesis pipeline followed by a linguistically guided re-engineering phase to ensure semantic variety and answer diversity. The entire dataset underwent two phases of domain expert validation by experienced botanists, ensuring clinical accuracy and relevancy to the field.
    
    \item \textbf{Benchmarking and Domain Adaptation}: We benchmark nine open-source vision–language models on PlantExpertVQA in a zero-shot setting and find that current frontier models perform poorly on the dataset. We further demonstrate, through parameter-efficient fine-tuning of a 2B-parameter model on a small fraction of the dataset, that PlantExpertVQA enables substantial improvements across all question categories, establishing its value as both a benchmark and a training resource.
\end{enumerate}
These contributions make PlantExpertVQA a useful and reliable dataset for research in plant disease diagnosis using visual question answering.

\section*{Methodology}

\subsection*{Multi-Source Image Collection and Preprocessing}

To build a comprehensive and robust foundation for PlantExpertVQA, we curated a diverse collection of 45 open-source plant disease datasets. The compiled corpus extends substantially beyond the widely used PlantVillage repository, which alone contributed approximately 55,000 laboratory-controlled images of 14 crops; the remaining 44 datasets contribute field-acquired imagery across additional crop species and disease conditions, expanding the final corpus to 38 unique crop species and 89 disease conditions. A critical inclusion criterion for these datasets was the presence of an explicit, well-curated hierarchical directory structure similar to the meticulous organization found in the widely used PlantVillage repository. In these chosen datasets, each directory systematically encodes the image's crop species and health status (e.g., \texttt{Tomato\_\_Late\_blight} or \texttt{Apple\_\_Healthy}). This structural consistency was vital for ensuring accurate, automated data generation in subsequent stages. All source datasets were released under permissive open licenses (Creative Commons or equivalent), and the full provenance of every image including its originating dataset, license, and citation is preserved in the released metadata; a complete listing of the 45 source datasets is provided in Supplementary Table~S1.

We implemented a systematic preprocessing pipeline across this multi-source compilation to guarantee standardization. We first eliminated exact and near-duplicate images using a combination of MD5 and Perceptual Hashing (pHash), applied both within and across source datasets to remove redundancy that arises when overlapping crop–disease pairs are independently distributed by multiple repositories. All images were then standardized to a uniform resolution. To prevent geometric distortion of crucial biological features, such as lesion shapes and leaf margins, we applied aspect-ratio-preserving padding rather than simple cropping or stretching, filling any remaining area with black pixels. Finally, we conducted automated technical audits to filter out low-quality samples and artifact biases. This included evaluating Laplacian variance to remove excessive blurriness, analyzing intensity histograms to discard over- or under-exposed images, and performing a background bias assessment using HSV color thresholding. This rigorous curation process resulted in a highly standardized, expert-chosen visual foundation of 150,841 high-quality images ready for QA generation.

\subsection*{Disease Knowledge Base Construction}

\begin{figure}[H]
    \centering
    \includegraphics[width=\linewidth]{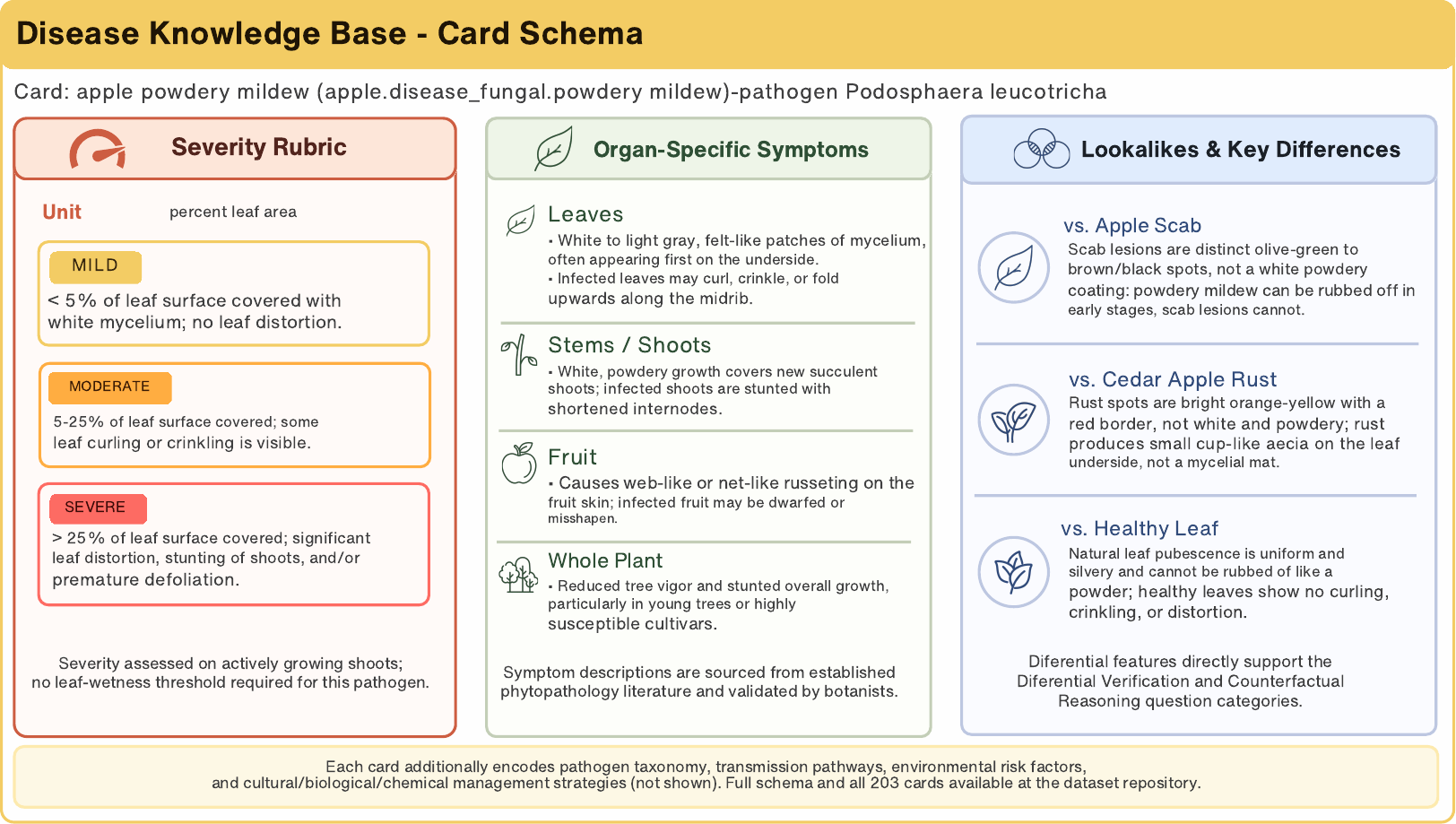}
    \caption{Structure of a Disease Knowledge Base card. Each of the 203 machine-readable cards encodes a specific crop--condition pair, detailing pathogens, symptoms, risk factors, and management. Illustrated here for apple powdery mildew are three key components: a severity rubric (left), organ-specific symptoms (centre), and lookalike conditions (right). This standardized schema directly grounds the generation of Visual Attribute, Severity, Differential, and Counterfactual questions.}
    \label{fig: kb card schema}
\end{figure}

To ground question and answer generation in established phytopathological science rather than in the priors of a general-purpose language model, we constructed a structured Disease Knowledge Base (KB) that encodes domain expertise in standardized json-schema. The KB consists of 203 cards, each corresponding to a single crop–condition pair drawn from the 38 crop species and 89 disease conditions present in the curated image collection; pairs include both healthy controls and pathological conditions of fungal, bacterial, viral, and abiotic etiology. The schema, illustrated in \autoref{fig: kb card schema}, was designed in collaboration with practising botanists and plant pathologists, and the substantive content of every card was extracted from established phytopathology literature \cite{agrios2005plant,strange2005plant} and validated through expert consensus before inclusion.

Each card encodes seven structurally distinct components. \textit{Pathogen taxonomy} records the kingdom-to-species classification of the causal agent, supporting the higher-order recall required by Specific Disease Identification questions. \textit{Transmission pathways} document vector species, dispersal mechanisms, and overwintering reservoirs, and \textit{environmental risk factors} record the temperature, humidity, and leaf-wetness conditions under which infection is favoured; together these components ground the answers to Causal Reasoning questions. The \textit{severity rubric} defines mild, moderate, and severe grades against quantitative thresholds (e.g., percent leaf area affected) and provides the discriminative basis for Severity Assessment questions. \textit{Organ-specific symptoms} catalogue canonical visual indicators separately for leaves, stems, fruit, roots, and the whole plant, providing the source descriptions for Visual Attribute Grounding questions. \textit{Lookalike conditions} record visually similar diseases together with explicit \textit{key differentiating features}, supporting Differential Verification and the negative cases of Counterfactual Reasoning. Finally, \textit{management strategies} catalogue cultural, biological, and chemical interventions, supporting Comprehensive Description and management-related question types. \autoref{fig: kb card schema} illustrates three of these components the severity rubric, organ-specific symptoms, and lookalikes with key differences for the apple powdery mildew card.

We deliberately encoded this domain knowledge in a structured format rather than as free-form text. The schema enforces uniformity across all 203 cards, and this uniformity carries over to the questions and answers generated from them. It also makes every answer traceable to a specific KB field, which supports both auditability and reproducibility. Finally, it lets the same card be queried along several cognitive axes (visual attributes, causal mechanisms, severity, and differential diagnosis) without authoring each axis separately. The full schema and the complete set of 203 cards are released alongside the dataset.

\subsection*{Programmatic QA Generation from the Knowledge Base}

Equipped with the structured Disease Knowledge Base described in the previous subsection, we generated the foundational pool of question--answer pairs by programmatically aligning each KB card with the images of the corresponding crop--condition pair. The hierarchical directory structure of the compiled image corpus, in which each image path explicitly encodes its crop and disease label, supplied the index linking each image to its KB card. For every (image, KB card) pair, our generation pipeline instantiated a fixed set of question templates, populating template slots with content drawn directly from the relevant KB fields: pathogen taxonomy, organ-specific symptoms, severity descriptors, lookalike comparisons, or management strategies, depending on the question category.


The question taxonomy comprises nine categories grouped into three levels of cognitive complexity. The categories range from foundational perception tasks such as Plant Species Identification, through detailed verification tasks such as Visual Attribute Grounding, to higher-order tasks such as Causal and Counterfactual Reasoning. The taxonomy was designed in consultation with three experienced graduate students from the Department of Botany, University of Dhaka, and was refined iteratively against early generation outputs. We deliberately avoided free-form generation by large language models at this stage: every question and every answer is traceable to an explicit KB field via a deterministic template, which eliminates hallucination risk and makes the entire foundational pool auditable and reproducible. The three levels and their constituent categories are shown in Supplementary Tables~S2, S3, and~S4 respectively (see Supplementary Information).

\subsection*{Data Refinement and Re-engineering}

Following the completion of programmatic generation, the foundational pool of 965,382 question--answer pairs was passed through a refinement pipeline whose purpose was twofold: to enrich the linguistic diversity of templated text without altering its meaning, and to remove non-verifiable or low-quality QA pairs through a combination of expert review and automated detection. Each stage of this pipeline is described in the subsections that follow.

\subsubsection*{Linguistic Diversification through Template-Based Paraphrasing}
In order to introduce a more diverse vocabulary, we applied Template-Based Paraphrasing. We first searched for the top repetitive questions and answers appearing more than 10,000 times. We rephrased the same text multiple times while preserving its original meaning. For questions, we first identified the most frequent templates. Next we used a pool of 10-15 high-quality paraphrasers to manually edit these templates. These annotators operated under strict guidelines to diversify the syntactic structure and phrasing without altering canonical pathological terminology, crop identifiers, or severity descriptors. This constraint was essential to prevent semantic drift, ensuring that while the linguistic variety of the questions increased, their underlying pathological accuracy remained deterministically anchored to the original Knowledge Base. By locking these core entities, we captured natural conversational diversity without compromising scientific precision. The resulting paraphrase pools thus expanded surface-level variety while keeping every question semantically faithful to its source template.

\begin{table}[H]
    \centering
    \small
    \begin{tabular}{cp{8cm}}
    \toprule
       \textbf{Original Question} & \textbf{Paraphrase Pool} \\
    \midrule
         & Identify the disease affecting this [Crop] leaf? \\
         & Can you diagnose the ailment present on this [Crop] foliage? \\
         What disease does this [Crop] leaf have? & What pathological condition is evident on this [Crop] leaf? \\
         & Which disease is indicated by the symptoms on this [Crop] leaf? \\
         & Please specify the disease observed on this [Crop] leaf. \\
    \bottomrule
    \end{tabular}
    \caption{Question and its Paraphrase Pool}
    \label{tab:Question Paraphrase Pool}
\end{table}

An example question and its paraphrase pool are shown in \autoref{tab:Question Paraphrase Pool}. At this stage, our team of specialists reviewed question variations from all nine categories to ensure scientific accuracy and consistency. We discarded all grammatically incorrect and excessively complex questions. Once validated, we replaced each question template with a randomly selected variation from its paraphrase pool. This process expanded our dataset vocabulary by 83.1\%, making the dataset more communicative and accessible.

\begin{figure}[H]
\centering
  \includegraphics[scale=.45]{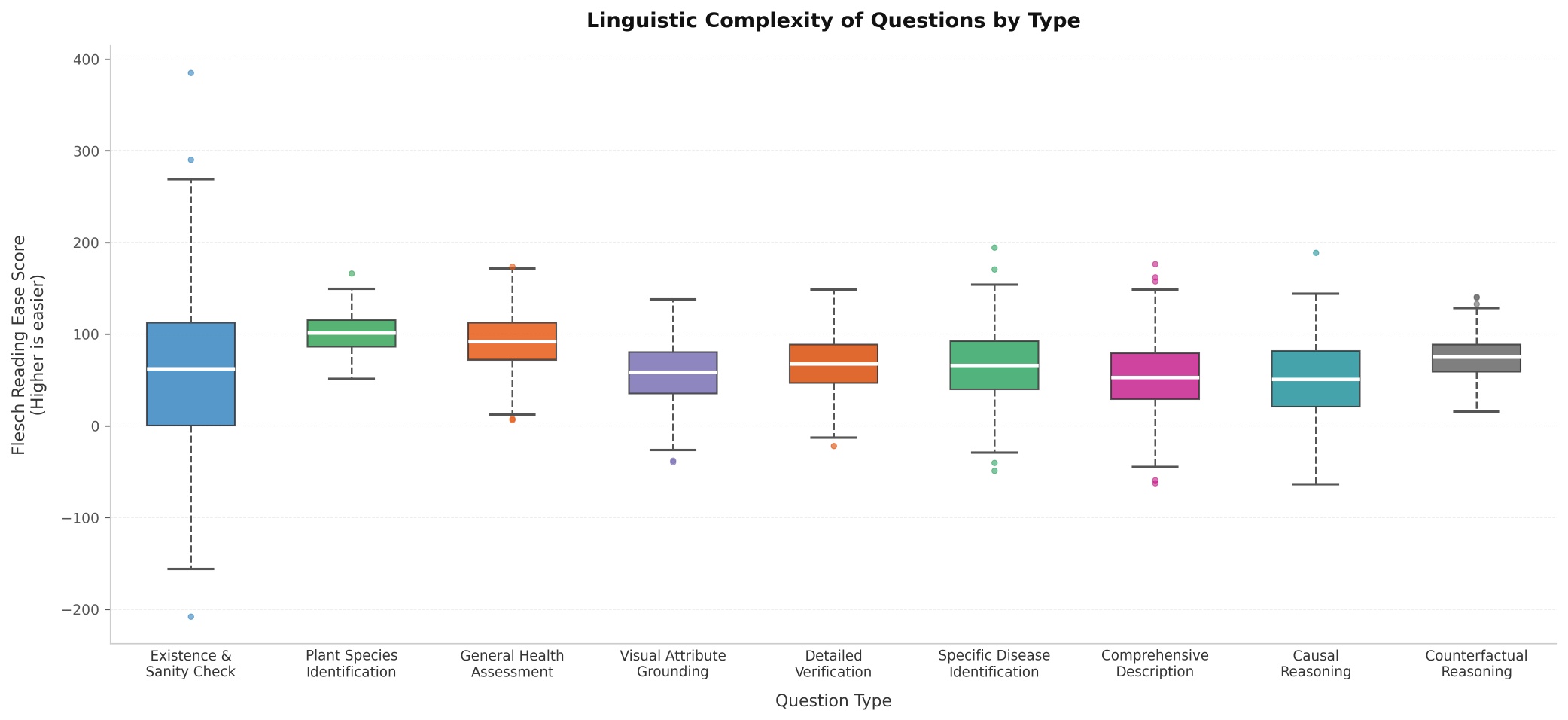}
  \caption{Question Linguistic Complexity}
  \label{fig:ques ling complx}
\end{figure}

We measured the comprehensibility of each question type through the Flesch Reading Ease Score. This score is calculated by considering average sentence length and the average number of syllables per word. Higher scores indicate better readability. \autoref{fig:ques ling complx} shows that this score varies across all question categories, indicating that the dataset covers both simple and complex inquiries. \\

\FloatBarrier
For answers, we employed a similar strategy. The dataset initially showed answer-side bias, in which a small fixed group of answers was provided for a given disease. To counter this, we first identified the ``answer bundles'' for each major disease. We then replaced them with a diverse pool of descriptive answers. For instance, all answers related to \texttt{Late\_blight} (e.g., \textbf{``This is a leaf with Late\_blight.'', ``The cause is Late\_blight.''}) were replaced by a random selection of more elaborative answers such as: \textbf{``The large, dark, water-soaked lesions are a key sign of Late Blight.''} and \textbf{``This is a classic presentation of Late Blight, caused by Phytophthora infestans.''}.

\begin{figure}[H]
    \centering
    \includegraphics[scale=0.4]{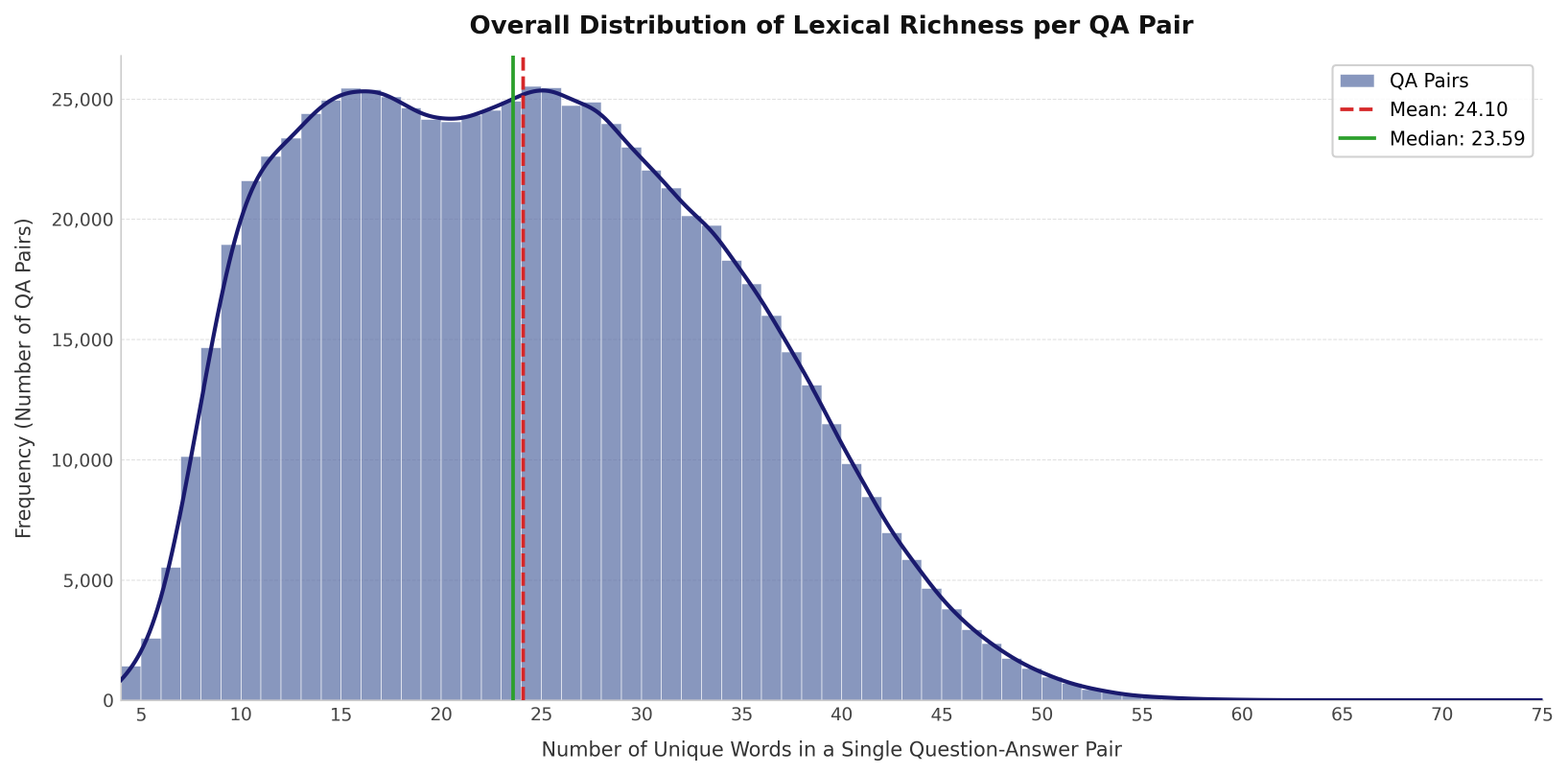}
    \caption{Lexical richness in QA pairs through re-engineering}
    \label{fig:lex rich qa pair}
\end{figure}

\begin{wraptable}{r}{0.55\textwidth}
\centering
\small
\begin{tabular}{lccc}
\toprule
\textbf{Metric} & \textbf{Initial} & \textbf{Final} & \textbf{Increase} \\
\midrule
Question Vocabulary & 1,789 & 3,275 & 1,486 \\
Answer Vocabulary   & 148   & 4,245 & 4,097 \\
\bottomrule
\end{tabular}
\caption{Vocabulary Growth in Question and Answer Sets}
\label{tab:vocab_growth}
\end{wraptable}

The number of unique words per QA pair was calculated through the Lexical Richness Score. \autoref{fig:lex rich qa pair} shows a mean count of 24.1 unique words per QA pair, meaning each pair uses approximately 24 words. This indicates that the generated data has a rich range of vocabulary while maintaining clarity.

\begin{wrapfigure}{r}{0.45\textwidth}
    \centering
    \includegraphics[width=0.43\textwidth]{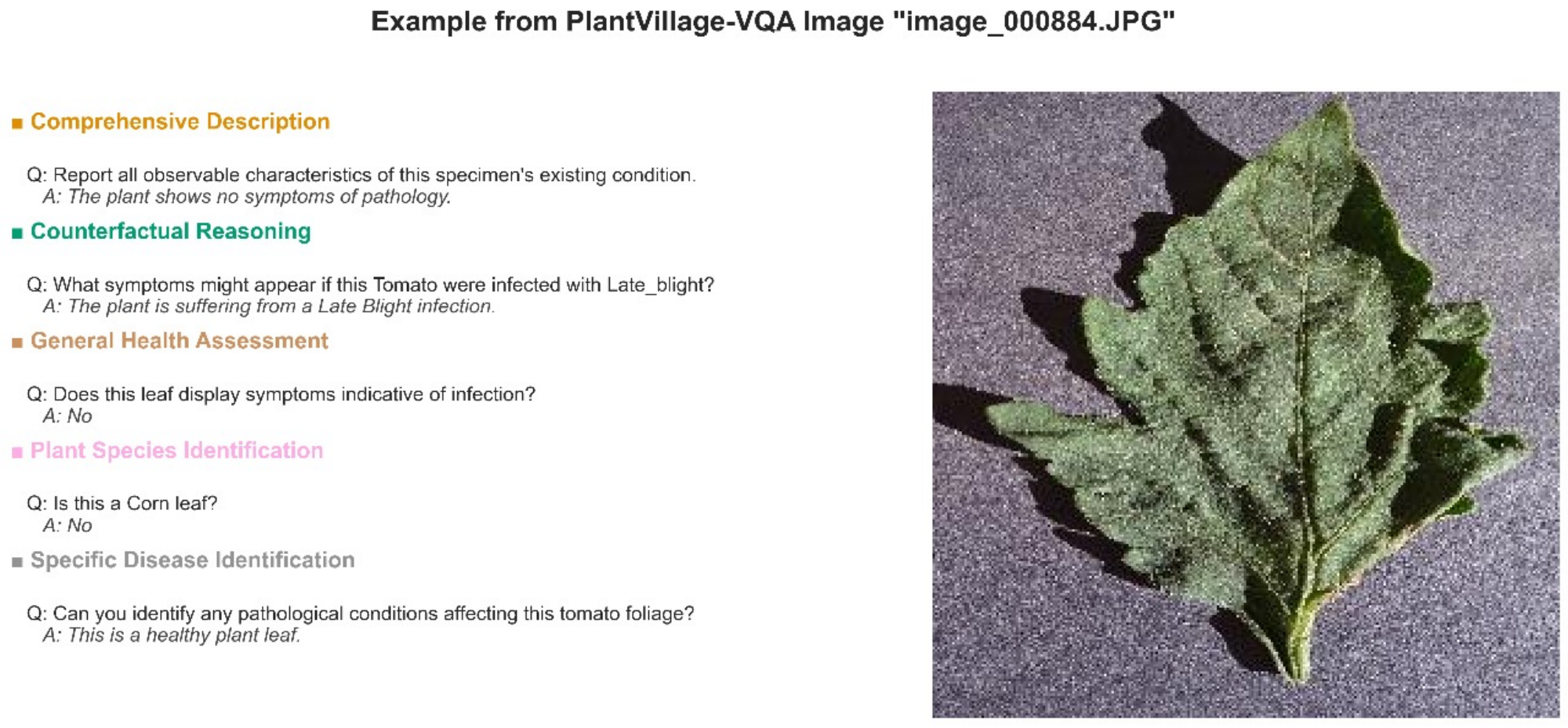}
    \caption{Image and its associated Questions}
    \label{fig:image_with_question}
\end{wrapfigure}

As a result the number of unique words in questions and answers increased as shown in \autoref{tab:vocab_growth}. An image template with the associated paraphrased QA pairs across multiple question categories is shown in \autoref{fig:image_with_question}.

\subsubsection*{Targeted Stratified Undersampling for Structural Balance}

Next we investigated if each question category was structurally balanced. We discovered that
primarily four types of questions: \textbf{Visual Attribute Grounding}, \textbf{Detailed
Verification}, \textbf{General Health Assessment} and \textbf{Plant Species Identification}
contributed to the overall imbalance.

\begin{figure}[H]
    \centering
    \includegraphics[scale=0.50]{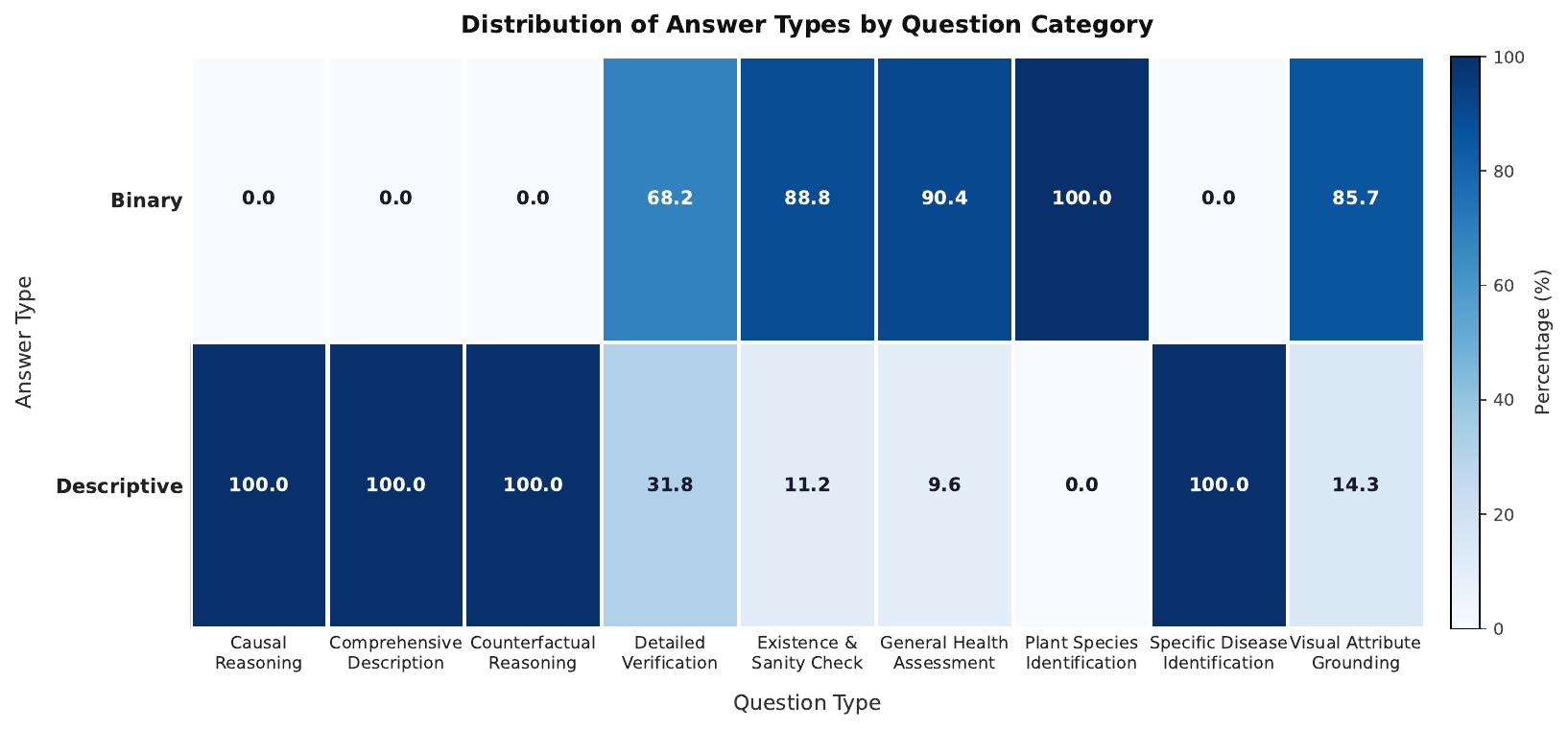}
    \caption{Heatmap of Answer type by Question Type}
    \label{fig: ans heat map}
\end{figure}

\autoref{fig: ans heat map} shows the heatmap of the binary and descriptive answers per
question category. It can be seen that all four aforementioned categories have above 80\%
binary answers. Further analysis showed that on average 62.6\% of these questions were skewed
toward negative answers. To solve this issue, we employed targeted stratified undersampling.
Instead of undersampling the entire dataset, we first extracted the QA pairs only from the
problematic question types. There, we retained 100\% of the ``Yes'' answer pairs and randomly
reduced the ``No'' answer pairs to create a balanced 40/60 binary ratio. The 40/60 ratio was
a heuristic choice, employed specifically to preserve most of the dataset while also reducing
the structural imbalance. Finally we merged the reduced QA bundles back into the original
dataset. This process removed \textbf{139,905} QA pairs from the structurally imbalanced
categories, resulting in an improved and balanced structure: the overall binary answer ratio
being 59.4\%. Although described here alongside the other refinement
operations, this stratified balancing was applied as the final step of corpus construction,
after the two expert-review phases; consequently the intermediate corpus of 905,182 QA pairs
reported in the Technical Validation precedes balancing, and the 139,905 pairs removed at this
stage are reflected in the final corpus total of 765,186 QA pairs.

\section*{Data Records}\label{sec:Data_Records}


The complete PlantExpertVQA dataset is publicly available on the Hugging Face Hub at \url{https://doi.org/10.57967/hf/9145}~\cite{sakib2025plantexpertvqadata} under a CC BY 4.0 licence. The deposit is organised into three components, the image corpus, the question--answer records, and the structured Disease Knowledge Base, together with the predefined train, validation and test partition described below.

\textbf{Images.} The 150,841 preprocessed images are provided in JPEG format. They are organised in a hierarchical directory structure in which each parent folder encodes the crop species and the health or disease status of its contents (e.g., \texttt{Tomato\_\_Late\_blight}, \texttt{Apple\_\_healthy}), mirroring the convention of the source repositories. Every image carries a unique \texttt{image\_id} that links it to the corresponding question--answer records and to its provenance metadata.\\

\textbf{Question--answer records.} The 765,186 QA pairs are released as tabular records (CSV, with an equivalent JSON Lines representation). Each record contains the following fields: \texttt{qa\_id} (a unique identifier for the QA pair); \texttt{image\_id} (the identifier of the grounding image); \texttt{crop} and \texttt{disease} (the crop species and condition); \texttt{severity} (the severity grade, where applicable); \texttt{question\_category} (one of the nine taxonomy categories); \texttt{cognitive\_level} (Level~1--3); \texttt{question\_text} and \texttt{answer\_text} (the natural-language QA pair); and \texttt{split} (train, validation, or test).

\textbf{Disease Knowledge Base.} The 203 crop--condition cards are provided as machine-readable JSON files, one per crop--condition pair. Each card exposes structurally distinct fields for pathogen taxonomy, transmission pathways, environmental risk factors, the severity rubric, organ-specific symptoms (catalogued separately for leaves, stems, fruit, roots, and the whole plant), lookalike conditions with their key differentiating features, and management strategies. Every answer in the QA records is traceable to a specific KB field, as detailed in the Methods.

\textbf{Data splits and provenance.} The corpus is partitioned at the image level into train (535,881 QA pairs; 70\%), validation (76,384; 10\%), and test (152,921; 20\%) subsets; the partition is image-disjoint and is recorded both in the \texttt{split} field and as separate split files. The full provenance of every image and its originating source dataset, licence, and bibliographic citation is preserved as per-image metadata; the 45 constituent source datasets are enumerated in Supplementary Table~S1.

\section*{Technical Validation}

To ensure dataset quality, we combined automated quality evaluation with two phases of expert review by experienced botanists. Given the extensive scale of the generated corpus, relying solely on manual inspection was logistically improbable, while purely automated metrics often fail to capture nuanced clinical inaccuracies. Therefore, we established a synergistic hybrid workflow: initial broad-scale human oversight identifies systemic patterns, which subsequently informs algorithmic outlier detection to pinpoint anomalies for a final, targeted manual review. The following sections describe each step of the validation pipeline.

\subsection*{Domain Expert Review: Phase One}
\begin{figure}[H]
    \centering
    \includegraphics[width=\linewidth]{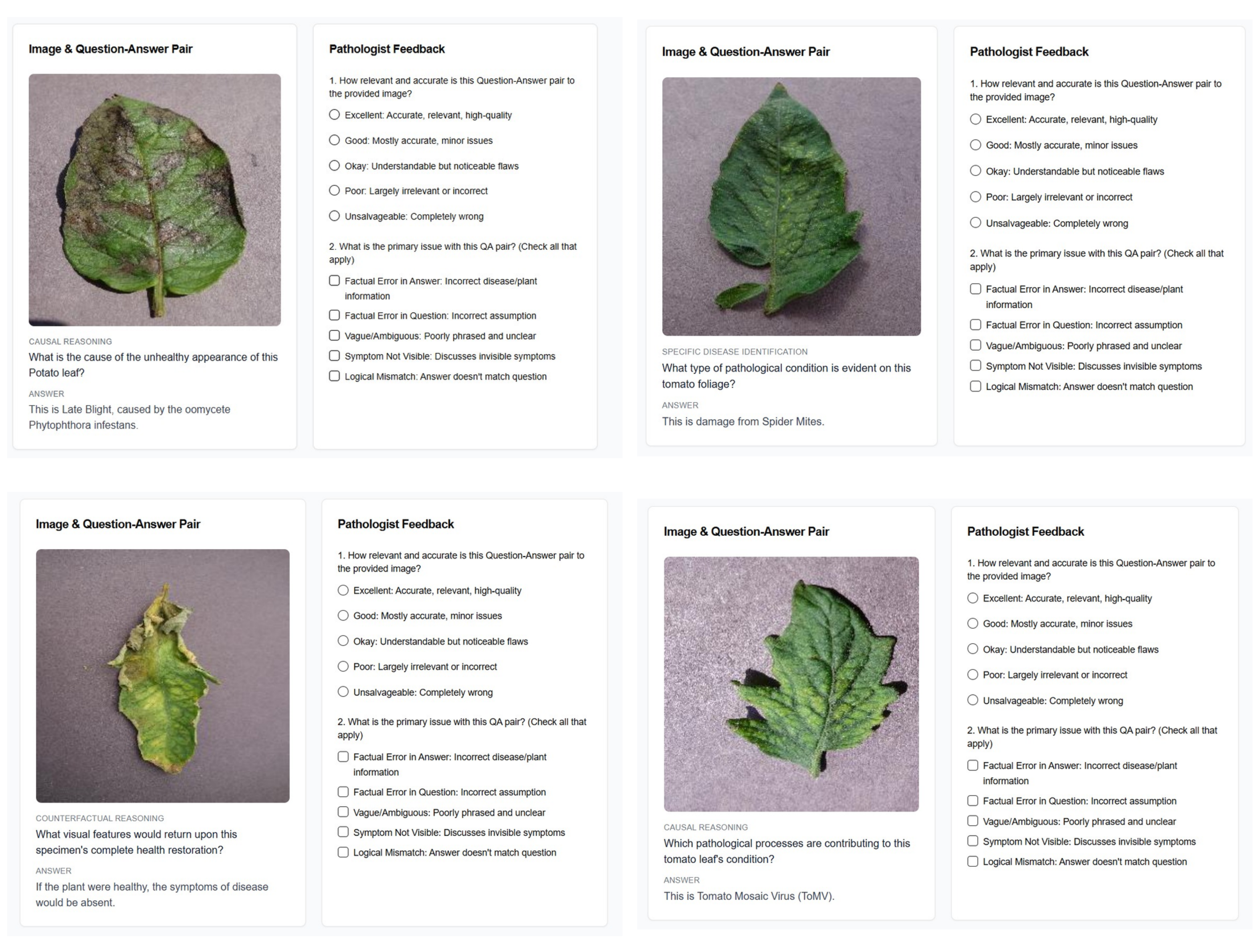}
    \caption{Phase One Expert Review Form}
    \label{fig: phase 1 form}
\end{figure}

The primary objective of this initial evaluation phase was to establish a baseline of scientific validity across the foundational question-answer pairs and intercept any recurring structural flaws. To facilitate this massive undertaking, we created a custom web interface hosting our entire dataset. This interface allowed a team of experienced botanists to efficiently review question-answer relevancy and identify related issues through a structured submission form. An example page from the website is shown in \autoref{fig: phase 1 form}.


Initial specialist feedback showed satisfactory performance across eight question categories. However, some associated answers relied heavily on generic fallback templates that contained no correlation with the questions. This was prevalent in the Counterfactual Reasoning category. While the questions themselves were correctly posed hypothetical scenarios (e.g., ``What visual features would be different if this plant were healthy?''), their answers were occasionally paired with simple diagnostic statements (e.g., A: ``The diagnosis is Tomato Yellow Leaf Curl Virus.'').
We identified this problem as poor specificity in counterfactual answer generation. To address it, we applied a Hierarchical Correction Pipeline to the 94,500 counterfactual QA pairs whose answers relied on generic fallback templates.

\subsubsection*{Implementation of the Hierarchical Correction Pipeline}
First, we leveraged the dataset's own internal knowledge to mine all 57,200 questions from the Visual Attribute Grounding category. These contained expert-phrased canonical descriptions of visual symptoms (e.g., ``Does the leaf exhibit dark, concentric `bullseye' rings?''). We programmatically extracted these descriptions and mapped them to key symptom words (e.g., `bullseye' $\to$ ``dark, concentric `bullseye' rings''), creating a \texttt{canonical\_phrase\_map}.

\begin{wraptable}{r}{0.55\textwidth}
    \centering
    \small
    \begin{tabular}{lc}
    \toprule
        \textbf{Category} & \textbf{Count} \\
    \midrule
        Initial Pool & 94,500 QA Pairs \\
        Corrected (Regenerated Pool) & 34,300 QA Pairs (36.3\%) \\
        Deleted (Non-Verifiable) & 60,200 QA Pairs (63.7\%) \\
    \bottomrule
    \end{tabular}
    \caption{Quantitative Outcome of Hierarchical Correction}
    \label{tab:quant outcome}
\end{wraptable}

To enhance counterfactual specificity we adopted a simple, defensible rule: fix when verifiable; delete when not. A QA pair was considered verifiable when it contained all three of the following: (i) valid \texttt{crop\_disease} provenance, (ii) a \texttt{disease\_keyword} that mapped to a condition, and (iii) an expert symptom phrase for that condition. If any element above was missing (e.g., a generic answer such as ``The plant is healthy.''), the QA pair was deleted. We then regenerated the valid counterfactual answers using the template: ``A healthy leaf would not show \ldots [canonical symptom] \ldots''. We refrained from using LLMs for question refinement to eliminate hallucination risk and preserve traceability. The numerical effect of the process on the counterfactual question category is shown in \autoref{tab:quant outcome}.

This ensured every surviving counterfactual answer was logically responsive, symptom-grounded, and reproducible from code and maps. A further 27,481 counterfactual pairs whose answers were deterministically grounded in the Disease Knowledge Base at generation time were exempt from this pipeline and merged with the 34,300 corrected pairs, yielding the 61,781 counterfactual pairs in the final corpus.

\subsubsection*{Impact of Logical Correction}
The Hierarchical Correction reduced the model's reliance on generic fallback answers, indicating a significant improvement in the counterfactual question category. \autoref{tab:comparison_before_after_specialist_review} provides a quantitative comparison of the dataset before and after final refinement.
\begin{table}[H]
\centering
\small
\setlength{\tabcolsep}{10pt}
\renewcommand{\arraystretch}{1.4}
\begin{tabular}{@{}lccc@{}}
\toprule
\textbf{Metric} 
  & \textbf{Before Feedback} 
  & \textbf{After Feedback} 
  & \textbf{Improvement} \\
\midrule
Generic answers (\% of counterfactuals) 
  & 55.03\% 
  & 5.49\% 
  & $\downarrow$ 90\% reduction \\[4pt]
Single generic template frequency 
  & 10{,}800 
  & 1{,}370 
  & $\downarrow$ 87\% reduction \\
\bottomrule
\end{tabular}
\caption{Comparison of quality metrics before and after specialist feedback and correction.}
\label{tab:comparison_before_after_specialist_review}
\end{table}

\subsection*{Automated Quality Evaluation and Outlier Detection}
After initial botanist review and correction, we conducted a thorough analysis of the entire database. We developed an automated pipeline to highlight suspicious QA pairs for expert re-evaluation. This process quantified abstract notions of quality, such as ``simplicity'', ``vagueness'' and ``relevance'' using established techniques from natural language processing and information theory.

\begin{figure}[H]
    \centering
    \includegraphics[width=\textwidth]{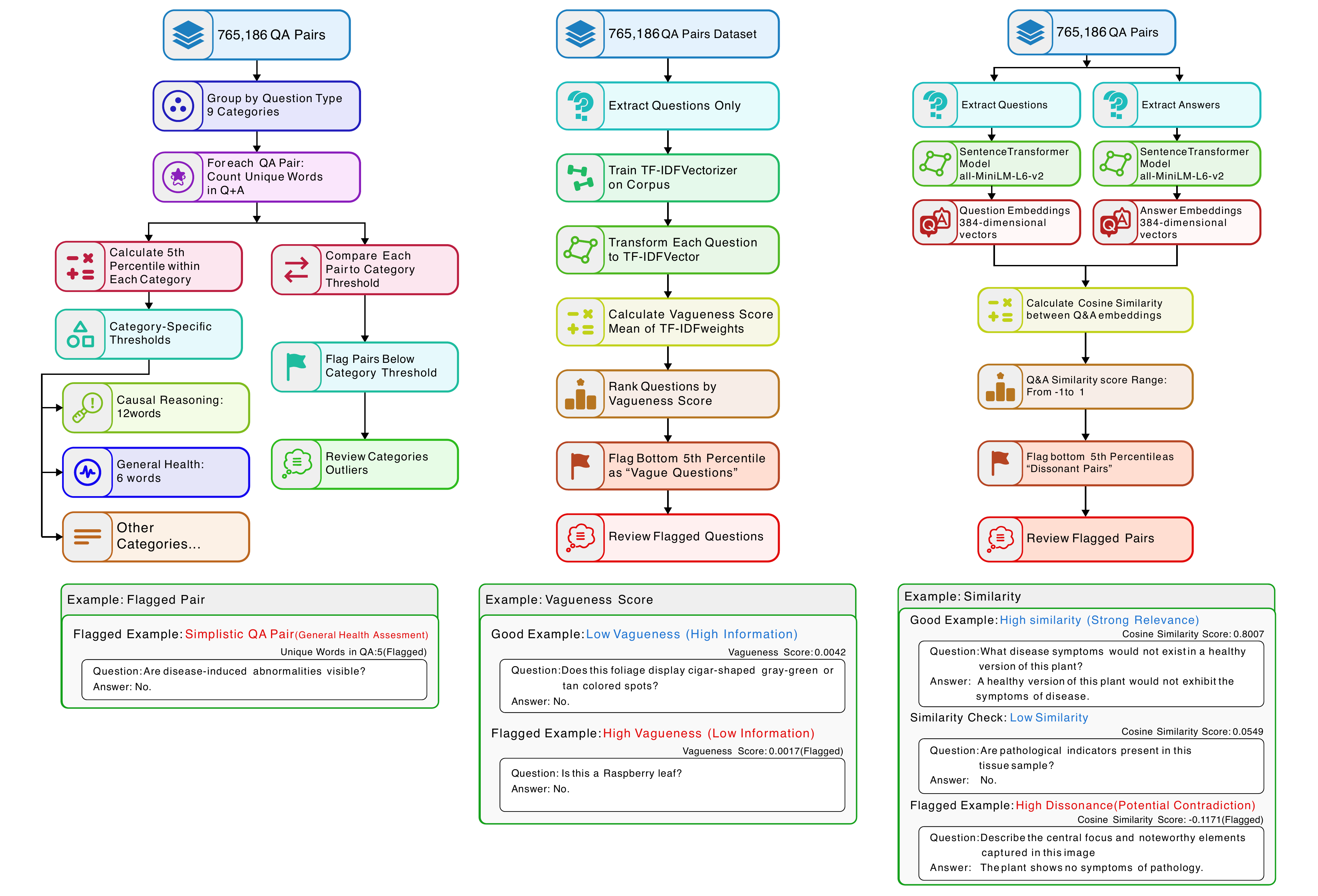}
    \caption{Automated quality evaluation and outlier detection workflow: relative simplicity (left), vagueness score assessment (center), and semantic dissonance (right).}
    \label{fig: automated outlier detection}
\end{figure}

\noindent
Three distinct analyses were performed on the post-correction corpus of 905,182 QA pairs as shown in \autoref{fig: automated outlier detection}. These metrics do not indicate inaccuracy in questions, but low conversational ability. For example, the question ``Is this a Raspberry leaf?'' is short but not vague in intent. These analyses allowed us to conduct a rigorous recheck of the entire database for communicative value.

\noindent
\textbf{Relative Simplicity:} We assessed relative simplicity by counting unique words per QA pair and comparing them with the mean unique-word counts of their question category. Questions below the 5th percentile threshold for each category (e.g., a Causal Reasoning question with 5 unique words) were flagged as outliers. The flowchart on the left of \autoref{fig: automated outlier detection} shows this assessment.

\noindent
\textbf{Vagueness Score:} We hypothesised that vague or low-value questions rely heavily on common words (e.g., ``what,'' ``is,'' ``leaf'') and lack specific keywords. To quantify this, we calculated a vagueness score using the TF-IDF weight of every word in a question. Lower scores indicated less informative content, and the bottom 5th percentile of such questions were flagged. The center flowchart in \autoref{fig: automated outlier detection} shows the process.

\noindent
\textbf{Semantic Similarity:} To detect semantic alignment between QA pairs, we used sentence-transformer embeddings (all-MiniLM-L6-v2) and computed QA similarity via cosine similarity. Lower scores identified weaker question-answer relevancy. This process is shown in the right flowchart of \autoref{fig: automated outlier detection}. In total, 118,600 QA pairs were flagged by the outlier detection pipeline.

\subsection*{Domain Expert Review: Phase Two}
\begin{wrapfigure}{r}{0.45\textwidth}
    \centering
    \includegraphics[width=0.43\textwidth]{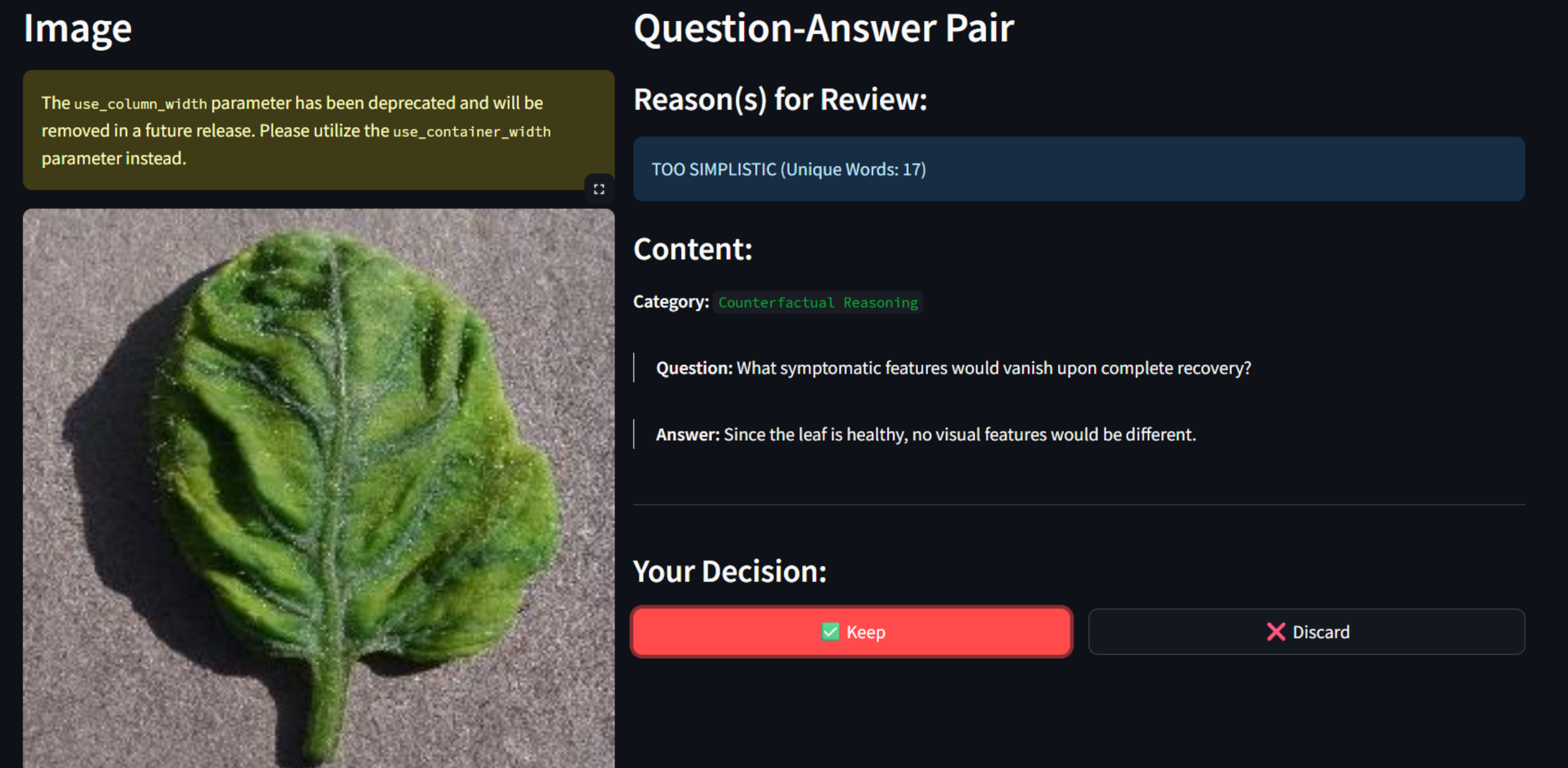}
    \caption{Phase Two Expert Review Form}
    \label{fig:Phase2form}
\end{wrapfigure}

In phase two, we focused only on the 118,600 QA pairs flagged by the automated outlier detection pipeline. We created a second web interface that displayed samples from this compilation. This form allowed reviewers to either keep or discard each QA pair. An example page of the form is shown in \autoref{fig:Phase2form}.

During this final inspection, our team of botanists reviewed a random sample of 2,837 flagged forms, of which only 91 were discarded. This corresponds to a sample retention rate of 96.8\%, indicating that the automated pipeline's flagging was conservative and that the surviving corpus was ready for model application and benchmarking. These two phases of expert validation, supported by the automated outlier detection pipeline, yielded the final PlantExpertVQA corpus of 765,186 high-quality QA pairs.

\FloatBarrier

\section*{Comparison with Existing Plant-Disease VQA Datasets}

To position PlantExpertVQA within the existing landscape of agricultural visual question answering resources, we compare it against the principal plant-disease VQA datasets reported in the recent literature. The comparator set comprises PlantVillageVQA, the v1 release that PlantExpertVQA extends; CDDM~\cite{Liu_2024}, the largest published plant-disease VQA dataset prior to this work; and CDwPK-VQA~\cite{zhao2023cdwpkvqa}, a knowledge-guided dataset constructed under a methodologically similar premise to ours but at smaller scale. \autoref{tab:dataset comparison} summarises the comparison along seven dimensions covering scale, coverage, source diversity, generation methodology, and expert validation.

\begin{table}[H]
\centering
\small
\renewcommand{\arraystretch}{1.3}
\setlength{\tabcolsep}{6pt}
\begin{adjustbox}{max width=\textwidth}
\begin{tabular}{|l|c|c|c|l|l|l|l|}
\hline
\textbf{Dataset} & \textbf{QA Pairs} & \textbf{Images} & \textbf{\thead{Crops /\\Cond.}} & \textbf{Answer Format} & \textbf{Reasoning Depth} & \textbf{KB Grounding} & \textbf{Expert Validation} \\
\hline
AgroBench~\cite{shinoda2025agrobench} & NR & NR & 203 / 682 & Constrained (MCQ) & Multi-task (7 topics) & $\times$ & Expert Agronomist \\
\hline
LeafNet~\cite{quoc2026leafnet} & 13,950 & 186,000 & 22 / 62 & Constrained (MCQ) & Visual Identification & $\times$ & Author-reported \\
\hline
AgMMU~\cite{gauba2025agmmu} & 1{,}492 / 5{,}460$^{\dagger}$ & 12,481 & Broad Ag. & MCQ \& Brief Open & Factual Dialogic & Partial (AgBase) & USDA-Verified \\
\hline
CDDM~\cite{Liu_2024} & $\sim$1{,}000{,}000 & 137,000 & 16 / 60 & Short Open (GPT-4 gen.) & Surface Visual & $\times$ & Author-reported \\
\hline
CDwPK-VQA~\cite{zhao2023cdwpkvqa} & 22,320 & 2,748 & 10 / 19 & Brief Open & Multi-attribute & Partial (prior knowledge) & Author-reported \\
\hline
WheatRust-VQA~\cite{nanavaty2024wheatrust} & NR & 1{,}800$^{\ddagger}$ & 1 / 4 & Fixed Templated & Symptom Identification & $\times$ & Author-reported \\
\hline
\rowcolor{gray!12}
\textbf{PlantExpertVQA (Ours)} & \textbf{765,186} & \textbf{150,841} & \textbf{38 / 89} & Open-ended (Multi-sentence) & Causal, Counterfactual, Severity & Extensive (203-Card KB) & Two-phase Botanist \\
\hline
\end{tabular}
\end{adjustbox}
\caption{\textbf{Comparison of PlantExpertVQA against existing plant-disease VQA datasets and benchmarks.} }
\label{tab:dataset comparison}
\end{table}

Consequently, PlantExpertVQA uniquely bridges the gap between the sheer scale of LLM-generated datasets and the reliability of knowledge-grounded resources. We note that recent vision-language systems like AgroGPT~\cite{awais2025agrogpt} and LLaVA-PlantDiag~\cite{sharma2024llava} are excluded from this direct comparison; as instruction-tuned models trained on LLM-generated data without disclosed expert validation protocols, they are not standalone benchmark datasets.

\section*{The Final PlantExpertVQA Corpus}

\begin{wraptable}{r}{0.55\textwidth}
\centering
\renewcommand{\arraystretch}{1.20}
\setlength{\tabcolsep}{6pt}
\adjustbox{max width=0.55\textwidth}{%
\begin{tabular}{l r}
\toprule
\multicolumn{2}{l}{\textbf{Scale}} \\
\midrule
Total QA Pairs & 765,186 \\
Total Images & 150,841 \\
Average QA Pairs per Image & 5.07 \\
\midrule
\multicolumn{2}{l}{\textbf{Coverage}} \\
\midrule
Unique Crop Species & 38 \\
Unique Disease Conditions & 89 \\
Unique Crop $\times$ Condition Combinations & 203 \\
\midrule
\multicolumn{2}{l}{\textbf{Question Categories}} \\
\midrule
Specific Disease Identification & 135{,}009 \enskip (17.64\%) \\
\rowcolor{gray!8}
Comprehensive Description       & 116{,}424 \enskip (15.22\%) \\
Plant Species Identification    & 108{,}447 \enskip (14.17\%) \\
\rowcolor{gray!8}
Causal Reasoning                &  \phantom{0}89{,}986 \enskip (11.76\%) \\
Detailed Verification           &  \phantom{0}78{,}829 \enskip (10.30\%) \\
\rowcolor{gray!8}
General Health Assessment       &  \phantom{0}66{,}572 \enskip (8.70\%)  \\
Counterfactual Reasoning        &  \phantom{0}61{,}781 \enskip (8.07\%)  \\
\rowcolor{gray!8}
Existence \& Sanity Check       &  \phantom{0}60{,}457 \enskip (7.90\%)  \\
Visual Attribute Grounding      &  \phantom{0}47{,}681 \enskip (6.23\%)  \\
\midrule
\multicolumn{2}{l}{\textbf{Splits} (image-level stratified)} \\
\midrule
Train      & 535{,}881 \enskip (70\%) \\
\rowcolor{gray!8}
Validation &  \phantom{0}76{,}384 \enskip (10\%) \\
Test       & 152{,}921 \enskip (20\%) \\
\bottomrule
\end{tabular}%
}
\caption{Composition of the final PlantExpertVQA corpus. Question-category counts follow the nine-category taxonomy of Supplementary Tables~S2--S4, aggregated post-refinement, and sum to the headline corpus total. Train, validation, and test splits are image-level stratified and image-disjoint, preserving crop and disease distributions across splits.}
\label{tab:dataset composition}
\end{wraptable}

Before proceeding to model evaluation, we summarise the composition of the released PlantExpertVQA corpus across four orthogonal axes: scale, coverage, question-category distribution, and the released splits. \autoref{tab:dataset composition} consolidates these statistics. With 765{,}186 expert-verified question--answer pairs grounded over 150{,}841 images, PlantExpertVQA averages 5.07 QA pairs per image, a ratio that supports multi-faceted evaluation of a single visual instance across cognitive levels rather than the one-question-per-image regime that characterises classification-style VQA datasets such as CDDM~\cite{Liu_2024}. Coverage extends across 38 crop species and 89 disease conditions, yielding 203 unique crop $\times$ condition combinations, the largest such cardinality reported for an expert-validated plant-disease VQA dataset to our knowledge.

The distribution of QA pairs across the nine-category taxonomy reflects the cognitive structure introduced in the Methodology. Higher-order reasoning categories occupy a substantial share of the corpus: Specific Disease Identification (17.64\%), Comprehensive Description (15.22\%), and Causal Reasoning (11.76\%) together account for over 44\% of all QA pairs, ensuring that the corpus is weighted toward the open-ended generative tasks that current vision-language models find most challenging~\cite{liu2023llava,dai2023instructblip}. Conversely, foundational perception categories such as Existence \& Sanity Check (7.90\%) and Visual Attribute Grounding (6.23\%) remain represented in sufficient volume to support discriminative pretraining and reliability auditing, but do not dominate the distribution as is common in classification-style benchmarks. This deliberate weighting follows the design principle articulated in Bloom-style hierarchical evaluation frameworks~\cite{gong2024bloomvqa}, in which higher-order cognitive tasks warrant proportionally greater representation when the explicit aim is to evaluate reasoning rather than perception alone.

The corpus is released with a 70/10/20 train, validation, and test partition stratified at the image level. Stratification is performed across question category, source dataset, crop species, and disease condition simultaneously, and the resulting splits are image-disjoint such that no image appears in more than one partition. This choice is deliberate: question-level stratification, which permits different QA pairs derived from the same image to fall into different splits, has been shown to overstate generalisation in multi-question-per-image VQA settings by leaking visual content across the train--test boundary. The image-level partition adopted here eliminates that risk and provides a conservative testbed against which subsequent vision-language models can be benchmarked and fine-tuned, as we demonstrate in the following section.

\section*{Evaluation and Benchmark}

\subsection*{Evaluation Metrics}

We evaluate model outputs using six complementary metrics that together capture lexical overlap, sequence-level alignment, and semantic similarity between generated answers and ground-truth references.

\textbf{Exact Match (EM)} measures the fraction of predictions that match the reference answer exactly after case normalisation:

\begin{equation}
    \text{EM} = \frac{1}{N} \sum_{i=1}^{N} \mathbf{1}\bigl[\hat{y}_i = y_i\bigr]
\end{equation}

where $\hat{y}_i$ is the model prediction, $y_i$ is the reference, and $N$ is the number of QA pairs. EM rewards precise short-form answers and is most informative for closed-set questions such as crop or disease identification.

\textbf{Token-F1} computes the harmonic mean of token-level precision and recall between the prediction and reference token sets, following the SQuAD evaluation convention \cite{rajpurkar2016squad}:

\begin{equation}
    \text{F1} = \frac{2 \cdot |\hat{T} \cap T|}{|\hat{T}| + |T|}
\end{equation}

where $\hat{T}$ and $T$ denote the multisets of tokens in the prediction and reference. Token-F1 relaxes the strict equality requirement of EM while still penalising both omissions and spurious tokens.

\textbf{BLEU-}$n$ \cite{papineni2002bleu} measures $n$-gram precision between the prediction and reference, with a brevity penalty to discourage degenerate short outputs:

\begin{equation}
    \text{BLEU-}n = \text{BP} \cdot \exp\!\left(\sum_{k=1}^{n} w_k \log p_k\right), \qquad \text{BP} = \min\!\left(1,\, e^{1 - r/c}\right)
\end{equation}

where $p_k$ is modified $n$-gram precision of order $k$, $w_k$ are uniform weights, $c$ is the prediction length, and $r$ is the reference length. We report BLEU-1 and BLEU-2 to capture vocabulary and short-phrase overlap.

\textbf{ROUGE-L} \cite{lin2004rouge} is computed as the F1-measure over the longest common subsequence (LCS) between prediction and reference:

\begin{equation}
    \text{ROUGE-L} = \frac{(1+\beta^2)\cdot R_{\text{LCS}} \cdot P_{\text{LCS}}}{R_{\text{LCS}} + \beta^2 \cdot P_{\text{LCS}}}, \qquad
    R_{\text{LCS}} = \frac{|\text{LCS}(\hat{y}, y)|}{|y|}, \qquad
    P_{\text{LCS}} = \frac{|\text{LCS}(\hat{y}, y)|}{|\hat{y}|}
\end{equation}

ROUGE-L captures sequence-level alignment without requiring contiguous matches, which makes it well suited to long-form, free-word-order pathology descriptions.

\textbf{BERTScore} \cite{zhang2020bertscore} computes semantic similarity through cosine alignment of contextual token embeddings produced by a pretrained language model. Given embeddings $\mathbf{x}_i$ for the reference and $\hat{\mathbf{x}}_j$ for the prediction:

\begin{equation}
    P_{\text{BERT}} = \frac{1}{|\hat{y}|} \sum_{\hat{x}_j \in \hat{y}} \max_{x_i \in y}\, \mathbf{x}_i^\top \hat{\mathbf{x}}_j, \qquad
    R_{\text{BERT}} = \frac{1}{|y|} \sum_{x_i \in y} \max_{\hat{x}_j \in \hat{y}}\, \mathbf{x}_i^\top \hat{\mathbf{x}}_j
\end{equation}

with $F_{\text{BERT}}$ defined as their harmonic mean. We adopt this metric to capture semantic equivalence between paraphrased outputs and references which is a property that purely lexical metrics cannot measure.

These metrics were selected to span the full evaluation spectrum: from strict surface-form agreement (EM, BLEU) through sequence-level structural alignment (ROUGE-L, Token-F1) to semantic equivalence (BERTScore). Together they support a comprehensive characterisation of model performance across the heterogeneous answer-length distributions present in PlantExpertVQA, ranging from single-word disease names to multi-sentence diagnostic descriptions. All metrics were computed on the complete test set, ensuring full coverage of the question-category, crop, and disease-condition distributions present in PlantExpertVQA.


\subsection*{Zero-Shot Benchmarking of Vision-Language Models}
\begin{table}[htbp]
\centering
\renewcommand{\arraystretch}{1.35}
\setlength{\tabcolsep}{6pt}
\adjustbox{max width=\textwidth}{%
\begin{tabular}{
    l
    r r r r r r
    r
}
\toprule
\multirow{2}{*}{\textbf{Model}}
  & \multicolumn{6}{c}{\textbf{Evaluation Metrics (\%)}}
  & \multirow{2}{*}{\textbf{\thead{Avg\\Len}}} \\
\cmidrule(lr){2-7}
  & \textbf{EM} & \textbf{Token-F1} & \textbf{BLEU-1} & \textbf{BLEU-2}
  & \textbf{ROUGE-L} & \textbf{BERTScore} & \\
\midrule
\rowcolor{gray!8}
Gemma-3-4B-IT          & 2.04  & 17.72          & 31.19          & 4.77 & \textbf{15.26} & \textbf{11.20} & 11.6 \\
Qwen3-VL-2B-Instruct   & 3.29  & \textbf{17.88} & \textbf{37.18} & \textbf{6.63} & 14.54 & 8.53  & 17.0 \\
\rowcolor{gray!8}
Qwen2-VL-2B-Instruct   & \textbf{8.91} & 13.76   & 36.57          & 3.23 & 13.16          & 7.95           & 3.2  \\
LLaVA-1.6 Mistral-7B   & 0.90  & 14.07          & 24.42          & 4.36 & 10.26          & 7.26           & 27.5 \\
\rowcolor{gray!8}
LLaVA-1.5-7B           & 7.39  & 10.21          & 26.87          & 0.58 & 9.87           & 3.15           & 2.5  \\
InstructBLIP Vicuna-7B  & 6.54  & 9.70           & 28.53          & 2.18 & 9.40           & $-$0.92        & 2.5  \\
\rowcolor{gray!8}
BLIP-2 FlanT5-XL        & 3.57  & 9.92           & 28.20          & 2.13 & 8.98           & $-$2.74        & 5.7  \\
CLIP ViT-L/14           & 3.31  & 7.44           & 15.30          & 2.27 & 6.95           & 3.48           & 7.1  \\
\rowcolor{gray!8}
BLIP VQA Large          & 5.43  & 6.65           & 21.41          & 0.04 & 6.64           & $-$1.59        & 1.3  \\
\bottomrule
\end{tabular}%
}

\caption{%
    \textbf{Zero-shot evaluation of nine vision--language models on PlantExpertVQA.}
    All metric values are percentages; Avg Len denotes average prediction length in tokens.
}
\label{tab:zero_shot}
\end{table}
We evaluated PlantExpertVQA against nine vision-language models spanning four principal architectural lineages: contrastive dual-encoders (CLIP \cite{radford2021clip}), classification-style VQA models (BLIP \cite{li2022blip}), Q-Former-bridged hybrids (BLIP-2 \cite{li2023blip2}, InstructBLIP \cite{dai2023instructblip}), and instruction-tuned multimodal LLMs (LLaVA-1.5/1.6 \cite{liu2023llava,liu2024llava}, Qwen-VL series \cite{bai2023qwenvl,bai2024qwen3vl}, Gemma-3 \cite{geminiteam2025gemma3}). 

Despite their pretraining scale, these frontier models struggle on PlantExpertVQA (\autoref{tab:zero_shot}). The strongest model, Gemma-3-4B-IT, attains only 15.26 ROUGE-L and 11.20 BERTScore-F1 against expert reference descriptions. Performance broadly scales with architectural modernity: dual-encoders and classification models cluster near 7 ROUGE-L, Q-Former hybrids and early MLLMs around 9--10, and recent instruction-tuned models reach 13--15. Consequently, a substantial gap to reference-quality pathology generation remains across all lineages.

Two key observations characterize this difficulty. First, Exact Match (EM) and ROUGE-L produce inverted rankings: Qwen2-VL leads in EM (8.91) by producing precise short answers, while Gemma-3 leads in ROUGE-L via longer descriptions, indicating no single architecture seamlessly handles both closed-set and long-form reasoning. Second, high BLEU-1 scores often pair with low or negative BERTScores (e.g., BLIP-2), demonstrating that superficial lexical overlap does not equate to semantic accuracy. Ultimately, mastering PlantExpertVQA requires domain-grounded, semantically precise generation which is a capability current open-source models lack without explicit domain adaptation.

\subsection*{Domain Adaptation through Parameter-Efficient Fine-Tuning}

To determine whether the zero-shot performance gap stems from fundamental architectural limits or missing domain knowledge, we conducted parameter-efficient fine-tuning on Qwen3-VL-2B-Instruct \cite{bai2024qwen3vl}. This model was selected for three reasons. First, it was the strongest 2B-class zero-shot baseline (14.54 ROUGE-L). This provides a robust starting point to unambiguously measure adaptation gains. Second, its architecture couples a language backbone to a Vision Transformer via a lightweight projector. This design represents the dominant paradigm in modern instruction-tuned models, which ensures our findings generalize. Third, its 2B parameter scale aligns with the computational and memory constraints of real-world agricultural decision-support systems. Such systems require edge-deployable efficiency.

\begin{table}[htbp]
\centering
\renewcommand{\arraystretch}{1.35}
\setlength{\tabcolsep}{5pt}
\adjustbox{max width=\textwidth}{%
\begin{tabular}{l r r r r r r r}
\toprule
\multirow{2}{*}{\textbf{Setting}}
  & \multicolumn{6}{c}{\textbf{Evaluation Metrics (\%)}}
  & \multirow{2}{*}{\textbf{\thead{Avg\\Len}}} \\
\cmidrule(lr){2-7}
  & \textbf{EM} & \textbf{Token-F1} & \textbf{BLEU-1} & \textbf{BLEU-2}
  & \textbf{ROUGE-L} & \textbf{BERTScore} & \\
\midrule
\rowcolor{gray!8}
 zero-shot (Best)
  & 2.04  & 17.72 & 31.19 & 4.77  & 15.26 & 11.20 & 11.6 \\
Qwen3-VL-2B-FT 
  & \textbf{18.85} & \textbf{64.65} & \textbf{66.18} & \textbf{41.74}
  & \textbf{61.70} & \textbf{63.27} & 27.2 \\
\bottomrule
\end{tabular}%
}
\caption{%
    \textbf{Performance of fine-tuned Qwen3-VL-2B against the strongest zero-shot performer on PlantExpertVQA.}
    All metric values are percentages; Avg Len denotes average prediction length in tokens.
    \textbf{Bold} marks the best result on each metric.
}
\label{tab:finetune_overall}
\end{table}

Adaptation was performed using Low-Rank Adaptation (LoRA) \cite{hu2022lora}, with rank-64 adapters applied to the query, key, value, and feedforward projection matrices of the language backbone. The vision encoder remained frozen, yielding 69.7 million trainable parameters -- 3.17\% of the model's 2.20 billion total. Training was conducted on a 500-image subset (2,337 QA pairs) of the PlantExpertVQA training split, stratified across question category, source, crop, and disease condition to mirror the evaluation distribution and verified to be image-disjoint from the test set. The deliberately small training set was designed to test whether the structured disease knowledge encoded in PlantExpertVQA provides sufficient supervision under limited data, rather than to maximise absolute performance. Optimisation used AdamW with a learning rate of $2 \times 10^{-4}$ and cosine decay over three epochs; the best checkpoint by validation loss was retained for evaluation.

\begin{table}[htbp]
\centering
\renewcommand{\arraystretch}{1.35}
\setlength{\tabcolsep}{8pt}
\adjustbox{max width=\textwidth}{%
\begin{tabular}{l r l r r}
\toprule
\textbf{Question Category}
  & \textbf{\thead{Best\\Zero-shot}}
  & \textbf{\thead{Leading\\Model}}
  & \textbf{\thead{Fine-tuned\\Qwen3-VL-2B}}
  & \textbf{Gain} \\
\midrule
\rowcolor{gray!8}
Plant Species Identification     & 39.68 & Qwen3-VL    & \textbf{57.56} & $+$17.88 \\
General Health Assessment        & 6.72  & BLIP-2      & \textbf{90.00} & $+$83.28 \\
\rowcolor{gray!8}
Visual Attribute Grounding       & 27.06 & Qwen3-VL    & \textbf{86.30} & $+$59.24 \\
Detailed Verification            & 24.32 & Qwen3-VL    & \textbf{62.67} & $+$38.35 \\
\rowcolor{gray!8}
Specific Disease Identification  &  7.10 & CLIP        & \textbf{57.96} & $+$50.86 \\
Comprehensive Description        & 10.43 & Qwen3-VL    & \textbf{63.57} & $+$53.14 \\
\rowcolor{gray!8}
Causal Reasoning                 & 21.64 & Qwen3-VL    & \textbf{74.48} & $+$52.84 \\
Counterfactual Reasoning         & 21.69 & Qwen3-VL    & \textbf{62.95} & $+$41.26 \\
\bottomrule
\end{tabular}%
}
\caption{%
    \textbf{Per-category ROUGE-L for fine-tuned Qwen3-VL-2B against the strongest zero-shot performer per category.}
    Results are reported across eight of the nine question categories defined in the PlantExpertVQA taxonomy; the Existence \& Sanity Check category is omitted as it serves a binary image-validity filter and is not separately evaluated during fine-tuning. The best zero-shot ROUGE-L across all nine evaluated models is reported for each category, with the leading model named in the adjacent column. \textbf{Bold} marks fine-tuned results; Gain $=$ Fine-tuned $-$ Best zero-shot. Values for Specific Disease Identification, Comprehensive Description, Causal Reasoning, and Counterfactual Reasoning are unweighted means over their constituent question sub-types.
}
\label{tab:finetune_category}
\end{table}

\autoref{tab:finetune_overall} contrasts the fine-tuned model against the strongest zero-shot performer on PlantExpertVQA, Gemma-3-4B-IT, which led the zero-shot benchmark on both ROUGE-L and BERTScore-F1 despite operating at twice the parameter count. The fine-tuned 2B model exceeds Gemma-3-4B-IT by $4.0\times$ on ROUGE-L (61.70 vs.\ 15.26), $5.7\times$ on BERTScore-F1 (63.27 vs.\ 11.20), and $9.2\times$ on exact match (18.85 vs.\ 2.04), with comparable margins across all remaining metrics. Average prediction length converges to 27.2 tokens, closely matching the reference distribution and indicating that the model has internalised the long-form pathology description style of PlantExpertVQA rather than continuing to default to short factual outputs.

Decomposed by question category in \autoref{tab:finetune_category}, the improvement is most pronounced on precisely the question types that resisted zero-shot adaptation across all evaluated systems: General Health Assessment rises from a best zero-shot value of 6.72 to 90.00 ROUGE-L, Visual Attribute Grounding from 27.06 to 86.30, Causal Reasoning from 21.64 to 74.48, and Comprehensive Description from 10.43 to 63.57. Two conclusions follow. First, the diagnostic-reasoning failures observed in zero-shot evaluation reflect a knowledge gap rather than a capability gap; once the structured pathological content of PlantExpertVQA is exposed to the model, it is acquired efficiently and produces marked downstream improvements. Second, PlantExpertVQA functions effectively both as a benchmark and as a training resource, parameter-efficient adaptation on a fraction of one percent of the training split suffices to elevate a compact 2B-parameter model above the strongest zero-shot models in our benchmark, including those at substantially larger parameter counts. We frame these results as a proof of concept for the utility of PlantExpertVQA in domain adaptation; comprehensive fine-tuning across additional architectures and at the full training scale is left to future work.

\subsection*{Comparison with Existing Plant-Disease VQA Models}

Existing plant-disease vision-language models report performance under widely heterogeneous evaluation protocols, which complicates direct head-to-head comparison. \autoref{tab:model comparison} summarises the reported performance of the principal comparator systems against our fine-tuned model.

\begin{table}[H]
\centering
\small
\begin{adjustbox}{max width=\textwidth}
\begin{tabular}{|l|l|l|l|c|}
\hline
\textbf{System} & \textbf{Test Set} & \textbf{Task Formulation} & \textbf{Reported Metric} & \textbf{Value} \\
\hline
\textbf{PlantExpertVQA Qwen3-VL-2B-FT (ours)} & PlantExpertVQA & Open-ended generation, 9 categories & ROUGE-L & \textbf{61.70} \\
 &  &  & BERTScore-F1 & \textbf{63.27} \\
 &  &  & Token-F1 & \textbf{64.65} \\
 &  &  & Exact Match & \textbf{18.85} \\
\hline
AgroGPT~\cite{awais2025agrogpt} & AgroEvals (6 tasks) & Multi-task agricultural QA & Task-level accuracy & Multi-task \\
\hline
LLaVA-PlantDiag~\cite{sharma2024llava} & PlantVillage-derived labels & Disease classification & Classification accuracy & $\sim$96.0 \\
\hline
CDDM Swin-T5~\cite{Liu_2024} & CDDM (292 unique answers) & Closed-set VQA & Classification accuracy & $\sim$99.94 \\
\hline
CDwPK-VQA~\cite{zhao2023cdwpkvqa} & CDwPK-VQA (22,320 pairs) & Closed-set VQA & Accuracy & 86.06 \\
\hline
\end{tabular}
\end{adjustbox}
\caption{Comparison of reported performance across plant-disease VQA models. Each system is evaluated on its own test distribution under its own metric, illustrating the methodological fragmentation in the field. Reported values for comparator systems are taken from the respective publications.}
\label{tab:model comparison}
\end{table}

Two observations follow from \autoref{tab:model comparison}. First, all comparator systems report classification-style accuracy against discrete answer sets, whereas PlantExpertVQA evaluates open-ended generation against multi-sentence pathology references; the two regimes measure substantively different capabilities and should not be ranked on a single axis. Second, the apparent absolute performance of comparator systems is partly an artefact of constrained answer spaces, most visibly in the case of CDDM, whose test set contains only 292 unique answers across approximately 1 million QA pairs. PlantExpertVQA addresses this fragmentation by providing a public unified benchmark spanning nine question categories with open-ended ground-truth references, supporting future evaluations in which competing systems can be retrained or zero-shot evaluated under identical conditions.

\section*{Limitations}

Several limitations of the present work warrant explicit acknowledgement. First, the fine-tuning experiment is intentionally scoped as a proof of concept: a single model architecture (Qwen3-VL-2B) was adapted on a small subset (2,337 QA pairs) rather than at full training scale, and comprehensive comparison across architectures and at the full training scale remains future work. Second, the dataset is monolingual, with all questions and answers in English, which limits direct applicability in agricultural contexts where local languages dominate; multilingual extension is a natural next step. Third, the source-dataset compilation, although diverse across 45 repositories, inherits the geographic and crop-distribution biases of its constituent datasets, with disease conditions affecting temperate and subtropical crops better represented than those of tropical and arid regions. Fourth, the Disease Knowledge Base, although extensive, is not exhaustive: 89 disease conditions cover the most prevalent pathologies in the source imagery but exclude rare, emerging, or geographically restricted conditions, and the static schema does not yet reflect ongoing taxonomic revisions or newly characterised pathogens. Finally, model performance is evaluated through automated lexical and semantic metrics; a complementary human evaluation by practising plant pathologists would provide further evidence of clinical utility and is left to future work.

\section*{Data Availability}

The PlantExpertVQA dataset, comprising all 150,841 preprocessed images, the 765,186 expert-verified question--answer pairs, and the 203-card Disease Knowledge Base, is publicly available on the Hugging Face Hub at \url{https://huggingface.co/datasets/SyedNazmusSakib/PlantExpertVQA}~\cite{sakib2025plantexpertvqadata} under a CC BY 4.0 licence. A detailed description of the constituent files, their formats, and their fields is provided in the Data Records section.

\section*{Code Availability}

The code for the programmatic QA generation pipeline, the data-refinement and template-paraphrasing steps, the automated outlier-detection pipeline, and the parameter-efficient fine-tuning experiments reported in this work is publicly available at \url{https://github.com/syed-nazmus-sakib/PlantExpertVQA}.

\section*{Acknowledgments}

We gratefully acknowledge the funding provided by the University Grants Commission of Bangladesh through the University of Dhaka during the fiscal year 2023-2024. We also acknowledge the University of Dhaka for providing the Article Processing Charge (APC) for this publication. We extend our sincere gratitude to the botanists from the Department of Botany, University of Dhaka for their invaluable support in conducting the expert review. Their contributions significantly enhanced the clinical accuracy and reliability of the question–answer validation process in the dataset. We would also like to acknowledge Ahnaf Tahmid Manan for his assistance with the graphical illustrations featured in this manuscript.

\section*{Author Contributions}

\textbf{Syed Nazmus Sakib}:  Conceptualization, Methodology, Investigation, Software, Visualization,  Writing - Review \& Editing.\\
\textbf{Nafiul Haque}: Methodology, Investigation, Visualization,  Writing - Original Draft.\\
\textbf{Mohammad Zabed Hossain}: Resources, Writing - Review \& Editing, Funding Acquisition. \\
\textbf{Shifat E. Arman}: Supervision, Funding Acquisition, Methodology, Resources,  Visualization,  Writing - Review \& Editing.

\section*{Competing Interests}

The authors declare that they have no known competing financial interests or personal relationships that could have appeared to influence the work reported in this paper.

\section*{Generative AI Usage}

The authors used a generative AI tool to help edit the language of this manuscript and improve its readability. It was not used to produce any of the data, results, or scientific claims. All text was checked by the authors, who take full responsibility for the final content.

\clearpage
\section*{Supplementary Information}
\setcounter{table}{0}
\renewcommand{\thetable}{S\arabic{table}}

\subsection*{S1. Constituent Open-Source Plant Disease Datasets}
\label{sec:supp_datasets}

The visual foundation of PlantExpertVQA is a compilation of 45 publicly released plant disease image datasets, enumerated in \autoref{tab:supp_datasets}. The compilation was assembled to extend substantially beyond the widely cited PlantVillage corpus~\cite{Hughes2015PlantVillage,mohanty2016using}, which despite its foundational role in the literature is restricted to 14 crop species captured under laboratory-controlled conditions. The remaining 44 datasets were drawn predominantly from recent Mendeley Data releases and contribute field-acquired imagery across additional crops, geographic origins, and pathological conditions, addressing the well-documented domain gap between controlled and in-field plant pathology imagery noted in prior work~\cite{singh2020plantdoc}. Each constituent dataset was admitted to the compilation on the basis of two criteria: (i) the presence of an explicit hierarchical directory structure in which the parent folder encodes both the crop species and the health or disease status of every image, enabling automated label-preserving integration with the structured Disease Knowledge Base described in the main text; and (ii) release under a permissive open licence (CC0 or CC BY 4.0) compatible with redistribution and derivative scientific use. The full provenance of every image in the released corpus, including its originating dataset, licence terms, and bibliographic citation, is preserved in the released metadata.

\renewcommand{\arraystretch}{1.25}
\setlength{\tabcolsep}{6pt}

\begin{small}
\begin{longtable}{@{} >{\centering\arraybackslash}p{0.5cm} p{6.5cm} p{3.5cm} p{2.0cm} p{1.8cm} @{}}
\caption{Constituent open-source plant disease image datasets aggregated to form the visual foundation of PlantExpertVQA. Each entry reports the dataset name as released by its authors, the primary crop or crop group covered, the licence under which the dataset was released, and the corresponding bibliographic reference.}
\label{tab:supp_datasets} \\
\toprule
\textbf{\#} & \textbf{Dataset Name} & \textbf{Primary Crop(s)} & \textbf{Licence} & \textbf{Ref.} \\
\midrule
\endfirsthead

\multicolumn{5}{l}{\textit{Table~\ref{tab:supp_datasets} continued from previous page}} \\
\toprule
\textbf{\#} & \textbf{Dataset Name} & \textbf{Primary Crop(s)} & \textbf{Licence} & \textbf{Ref.} \\
\midrule
\endhead

\midrule
\multicolumn{5}{r}{\textit{Continued on next page}} \\
\endfoot

\bottomrule
\endlastfoot

1  & PlantVillage                                                              & Multi-crop (14)         & CC0          & \cite{Hughes2015PlantVillage} \\
2  & Plant Pathology Challenge 2020                                            & Apple                   & CC BY 4.0    & \cite{PlantPathology2020} \\
3  & Apple Leaf Diseases (ICAR-CITH)                                           & Apple                   & CC BY 4.0    & \cite{AppleLeafDiseasesICAR2024} \\
4  & Chili Plant Leaf Disease and Growth Stage                                 & Chili                   & CC BY 4.0    & \cite{nirob2025chili} \\
5  & Banana and Banana Leaf Dataset                                            & Banana                  & CC BY 4.0    & \cite{das2025banana} \\
6  & CAIR-BGD-2025: Bottle Gourd Disease and Growth Stages                     & Bottle gourd            & CC BY 4.0    & \cite{nirob2025cairbgd} \\
7  & Plant Leaf Disease Recognition Dataset                                    & Multi-crop              & CC BY 4.0    & \cite{ahmed2024plantleaf} \\
8  & Papaya Leaf Disease Image Dataset                                         & Papaya                  & CC BY 4.0    & \cite{albanna2024papaya} \\
9  & Real-World Papaya Leaf (Bangladeshi Orchards)                             & Papaya                  & CC BY 4.0    & \cite{rashid2024papaya} \\
10 & Eggplant Leaf Disease Dataset                                             & Eggplant                & CC BY 4.0    & \cite{rafe2025eggplant} \\
11 & Eggplant Comprehensive Dataset                                            & Eggplant                & CC BY 4.0    & \cite{nirob2024eggplant} \\
12 & High-Resolution Eggplant Leaf Dataset                                     & Eggplant                & CC BY 4.0    & \cite{hasan2025eggplanthighres} \\
13 & BrinjalFruitX                                                             & Brinjal (eggplant)      & CC BY 4.0    & \cite{hasan2025brinjalfruitx} \\
14 & High-Resolution Lychee Plant Diseases                                     & Lychee                  & CC BY 4.0    & \cite{hossain2025lychee} \\
15 & BDLitchi                                                                  & Litchi                  & CC BY 4.0    & \cite{ahamed2025bdlitchi} \\
16 & LitchiLeaf4001                                                            & Litchi                  & CC BY 4.0    & \cite{hasan2025litchileaf4001} \\
17 & TLD-BD: Tea Leaf Dataset                                                  & Tea                     & CC BY 4.0    & \cite{ahmed2025tldbd} \\
18 & Advanced Tea Crop Disease Study                                           & Tea                     & CC BY 4.0    & \cite{ahmad2024advancedtea} \\
19 & Hibiscus and Tea Leaf Dataset                                             & Hibiscus, Tea           & CC BY 4.0    & \cite{billah2025hibiscustea} \\
20 & Hibiscus Leaf Diseases Classification                                     & Hibiscus                & CC BY 4.0    & \cite{billah2025hibiscus} \\
21 & Multi-crop Leaf Disease (Bitter gourd, Okra, Pumpkin, Ridge gourd)        & Multi-crop              & CC BY 4.0    & \cite{islam2025multicrop} \\
22 & SAR-CLD-2024: Cotton Leaf Disease                                         & Cotton                  & CC BY 4.0    & \cite{bishshash2024cotton} \\
23 & Cotton Leaf Image Dataset for Disease Classification                      & Cotton                  & CC BY 4.0    & \cite{ripon2025cotton} \\
24 & MoringaLeafNet                                                            & Moringa                 & CC BY 4.0    & \cite{khan2025moringaleafnet} \\
25 & MangoLeafBD                                                               & Mango                   & CC BY 4.0    & \cite{MangoLeaf2024} \\
26 & Comprehensive Mango Leaf Disease Dataset                                  & Mango                   & CC BY 4.0    & \cite{hossan2025mango} \\
27 & Sunflower Fruits and Leaves                                               & Sunflower               & CC BY 4.0    & \cite{rajbongshi2022sunflower} \\
28 & Sunflower Plant Health and Growth Stage                                   & Sunflower               & CC BY 4.0    & \cite{sagor2025sunflowergrowth} \\
29 & Multifaceted Rose Leaf Disease                                            & Rose                    & CC BY 4.0    & \cite{ahmad2025multifacetedrose} \\
30 & Tomato Disease Dataset                                                    & Tomato                  & CC BY 4.0    & \cite{liu2025tomato} \\
31 & Tomato Leaf Diseases                                                      & Tomato                  & CC BY 4.0    & \cite{hossain2025tomatoleaf} \\
32 & Tomato Leaf (Multiclass Detection and Classification)                     & Tomato                  & CC BY 4.0    & \cite{imtiaz2024tomato} \\
33 & Tomato (Bangladesh High-Resolution)                                       & Tomato                  & CC BY 4.0    & \cite{bapari2025tomato} \\
34 & AJLCD-2025: Arabian Jasmine Leaf Condition                                & Arabian jasmine         & CC BY 4.0    & \cite{ayon2025ajlcd} \\
35 & Durian Leaf Diseases (Vietnam)                                            & Durian                  & CC BY 4.0    & \cite{truong2025durian} \\
36 & BDRubberLeaf                                                              & Rubber                  & CC BY 4.0    & \cite{debnath2025bdrubberleaf} \\
37 & Cauliflower Diseases Identification                                       & Cauliflower             & CC BY 4.0    & \cite{matin2025cauliflower} \\
38 & CitrusLeafVision (Lemon)                                                  & Lemon                   & CC BY 4.0    & \cite{debnath2025citrusleafvision} \\
39 & Jackfruit Plant Leaf Disease                                              & Jackfruit               & CC BY 4.0    & \cite{huq2025jackfruit} \\
40 & AgriLeafNet (Fruit Tree Leaves)                                           & Multi-fruit             & CC BY 4.0    & \cite{haque2025agrileafnet} \\
41 & Pisum sativum (Pea) Image Dataset                                         & Pea                     & CC BY 4.0    & \cite{thite2025pisum} \\
42 & Turmeric Plant Leaf Disease                                               & Turmeric                & CC BY 4.0    & \cite{hossain2025turmeric} \\
43 & NGLD: Niphad Grape Leaf Disease                                           & Grape                   & CC BY 4.0    & \cite{dharrao2025ngld} \\
44 & Burmese Grape Leaf Disease                                                & Burmese grape           & CC BY 4.0    & \cite{rahman2025burmesegrape} \\
45 & UGV: Guava Leaves Disease Bangladesh                                      & Guava                   & CC BY 4.0    & \cite{hassan2025guava} \\

\end{longtable}
\end{small}

In aggregate, the 45 constituent datasets contribute approximately 321{,}000 raw images, of which 150{,}841 images survived the standardisation, de-duplication, and quality-audit pipeline described in the main text (Methodology, \textit{Multi-Source Image Collection and Preprocessing}). After consolidation against the Disease Knowledge Base, the surviving corpus spans 38 unique crop species and 89 distinct disease conditions, with overlapping crop--condition pairs across multiple source datasets resolved through perceptual-hash-based cross-dataset de-duplication. The dominant geographic origin of the compilation is South Asia, with a substantial concentration of datasets curated in Bangladesh, India, and Vietnam; this regional emphasis is acknowledged as a limitation in the main text and represents a target for future expansion towards under-represented tropical and arid agro-ecological zones. We note that the heavy reliance on recent Mendeley Data releases reflects a broader trend in agricultural computer vision towards open-access, field-acquired imagery~\cite{singh2020plantdoc}, and we anticipate that the compilation strategy adopted here will remain extensible as further such datasets become available.


\subsection*{S2. Detailed Question Category Taxonomy}
\label{app:qeustion_cat_detailed}

This section provides the detailed specification of the nine-category question taxonomy referenced in the Methodology. The taxonomy is organised into three levels of cognitive complexity, following the design principle articulated in Bloom-style hierarchical evaluation frameworks for multimodal comprehension~\cite{gong2024bloomvqa}: foundational perception and identification (Level~1), detailed analysis and verification (Level~2), and higher-order reasoning and inference (Level~3). For each category we report (i) its diagnostic purpose, (ii) the Disease Knowledge Base (KB) field or fields from which template slots are instantiated, and (iii) two to three representative templates drawn from the released generation script. Templates are shown in their canonical pre-paraphrase form; the linguistically diversified surface forms produced by the re-engineering pipeline are released alongside the dataset.

\subsubsection*{S2.1 Level~1: Foundational Perception and Identification}

Level~1 questions assess the model's capacity to identify discrete elements in the image with factual, closed-set answers. They establish baseline visual competence and act as sanity filters: a model that fails Level~1 questions cannot reasonably be expected to succeed at higher cognitive levels. The three categories are summarised in \autoref{tab: qtype level 1}.

\begin{table}[H]
\centering
\small
\begin{adjustbox}{max width=\textwidth}
\begin{tabular}{|L{3.5cm}|L{6cm}|L{7cm}|}
\hline
\textbf{Question Category} & \textbf{Purpose and KB Grounding} & \textbf{Representative Templates} \\
\hline

\textbf{1. Existence \& Sanity Check} &
This category filters out irrelevant or background images and verifies that the model recognises plant material as the primary subject. Templates are content-independent and do not require KB instantiation. &
\textit{What is the primary subject of this image?} \par
\textit{Does this image depict plant material?} \par
\textit{Is this a photograph of a plant or a plant part?} \\

\hline

\textbf{2. Plant Species Identification} &
For species-specific diseases, identifying the host plant is a prerequisite for downstream diagnosis. This category tests fine-grained visual categorisation. Templates are instantiated using the \texttt{crop.common\_name} field of the corresponding KB card, with both positive and negative variants generated to produce a balanced binary task. &
\textit{Is this a \mbox{[Crop]} leaf?} \par
\textit{Which crop species is depicted in this image?} \par
\textit{Can you identify the host plant shown here?} \\

\hline

\textbf{3. General Health Assessment} &
This category tests the model's ability to make a high-level binary judgement of the plant's overall health status. Templates are instantiated from the \texttt{issue\_type} field of the KB card, which distinguishes diseased from healthy entries. &
\textit{Is the plant in this image healthy?} \par
\textit{Does this leaf show any signs of disease or stress?} \par
\textit{Would you classify the depicted specimen as healthy or diseased?} \\

\hline
\end{tabular}
\end{adjustbox}
\caption{Question categories for Level~1 (Foundational Perception and Identification). KB fields refer to the schema introduced in the Methodology.}
\label{tab: qtype level 1}
\end{table}

\subsubsection*{S2.2 Level~2: Detailed Analysis and Verification}

Level~2 questions move from coarse identification to symptom-level analysis. They constitute the diagnostic core of PlantExpertVQA: the model must detect, and ground its responses in, the specific visual symptoms catalogued in the phytopathology literature~\cite{agrios2005plant,strange2005plant}. The two categories are summarised in \autoref{tab: qtype level 2}.

\begin{table}[H]
\centering
\small
\begin{adjustbox}{max width=\textwidth}
\begin{tabular}{|L{3.5cm}|L{6cm}|L{7cm}|}
\hline
\textbf{Question Category} & \textbf{Purpose and KB Grounding} & \textbf{Representative Templates} \\
\hline

\textbf{1. Visual Attribute Grounding} &
This category tests whether the model can detect specific, often subtle, visual symptoms that are referenced in a textual phrase. Templates are instantiated from the \texttt{symptoms} fields of the KB card, which catalogue canonical visual indicators separately for leaves, stems, fruit, roots, and the whole plant. The underlying symptom descriptions are sourced from established phytopathology literature \cite{agrios2005plant,strange2005plant} and validated through expert consensus during KB construction. &
\textit{Does the leaf exhibit dark, concentric ``bullseye'' rings?} \par
\textit{Are yellow halos visible around the lesions on this leaf?} \par
\textit{Is there a powdery white coating on the leaf surface?} \\

\hline

\textbf{2. Detailed Verification} &
This category combines species identification and disease identification into a single compositional binary query. Templates jointly populate \texttt{crop.common\_name} and \texttt{condition.common\_name} from the KB card. Both positive (correct crop and disease) and negative (mismatched crop, mismatched disease, or both) variants are generated. &
\textit{Is this \mbox{[Crop]} leaf infected with \mbox{[Disease\_Name]}?} \par
\textit{Does this image show a \mbox{[Crop]} plant affected by \mbox{[Disease\_Name]}?} \par
\textit{Can \mbox{[Disease\_Name]} be diagnosed from the symptoms visible on this \mbox{[Crop]} leaf?} \\

\hline
\end{tabular}
\end{adjustbox}
\caption{Question categories for Level~2 (Detailed Analysis and Verification). KB fields refer to the schema introduced in the Methodology.}
\label{tab: qtype level 2}
\end{table}

\subsubsection*{S2.3 Level~3: Higher-Order Reasoning and Inference}

Level~3 questions require the model to synthesise information across multiple visual cues, infer relationships between symptoms and underlying mechanisms, and reason about hypothetical or counterfactual scenarios. These are the open-ended generative tasks on which current vision-language models exhibit the largest performance gaps in our zero-shot evaluation. The four categories are summarised in \autoref{tab: qtype level 3}.

\begin{table}[H]
\centering
\small
\begin{adjustbox}{max width=\textwidth}
\begin{tabular}{|L{3.5cm}|L{6cm}|L{7cm}|}
\hline
\textbf{Question Category} & \textbf{Purpose and KB Grounding} & \textbf{Representative Templates} \\
\hline

\textbf{1. Specific Disease Identification} &
This is the direct open-ended diagnostic task. Unlike Detailed Verification, the model is not given a candidate disease; it must recall and provide the correct one from its internal knowledge. Answers are sourced from the \texttt{condition.common\_name} field of the KB card. &
\textit{Please provide a diagnosis for the condition shown.} \par
\textit{What specific disease is affecting this plant?} \par
\textit{Identify the pathological condition visible on this leaf.} \\

\hline

\textbf{2. Comprehensive Description} &
This category tests the model's ability to generate holistic summary descriptions that span multiple aspects of the image. Answers are composed by aggregating multiple KB fields (pathogen, key organ-specific symptoms, and severity grade) into a single descriptive paragraph. &
\textit{Provide a full description of the plant and its condition.} \par
\textit{Describe the plant, the visible symptoms, and the likely diagnosis.} \par
\textit{Summarise the visual evidence and provide a clinical assessment.} \\

\hline

\textbf{3. Causal Reasoning} &
This category probes whether the model can infer the causal agent and the conditions responsible for the visible symptoms. Answers are instantiated from the \texttt{condition.pathogen}, \texttt{transmission}, and \texttt{environmental\_risk\_factors} fields of the KB card. &
\textit{What is the cause of the unhealthy appearance of this leaf?} \par
\textit{Which pathogen is responsible for the symptoms shown?} \par
\textit{Under what environmental conditions does this disease typically develop?} \\

\hline

\textbf{4. Counterfactual Reasoning} &
This category tests the model's capacity to reason about a hypothetical state that is contrary to the visual evidence. Answers are constructed by negating the symptom descriptions of the affected KB card, e.g., ``a healthy leaf would not show \ldots[canonical symptom]\ldots'', ensuring counterfactual answers are grounded in the same KB field that drives Visual Attribute Grounding. This shared grounding underpins the Hierarchical Correction Pipeline described in the main text. &
\textit{If this plant were healthy, what visual features would be different?} \par
\textit{Which of the visible features would be absent in a healthy specimen?} \par
\textit{What changes would indicate that this plant has recovered from infection?} \\

\hline
\end{tabular}
\end{adjustbox}
\caption{Question categories for Level~3 (Higher-Order Reasoning and Inference). KB fields refer to the schema introduced in the Methodology.}
\label{tab: qtype level 3}
\end{table}

Across the three levels, every template in the released generation pipeline is mapped deterministically to a specific KB field or combination of fields. This design choice eliminates the hallucination risk associated with free-form generation by large language models~\cite{Liu_2024} and renders the entire foundational pool of 965{,}382 QA pairs auditable and reproducible from the released KB and template scripts. The post-refinement category counts and their proportions within the final 765{,}186-pair corpus are reported in the corpus-composition table of the main manuscript (section ``The Final PlantExpertVQA Corpus'').


\begin{thebibliography}{99}

\bibitem{bhuiyan2023bananasqueezenet}
Md Abdullahil Baki Bhuiyan, Hasan Muhammad Abdullah, Shifat E Arman, Sayed Saminur Rahman, and Kaies Al Mahmud.
\newblock BananaSqueezeNet: A very fast, lightweight convolutional neural network for the diagnosis of three prominent banana leaf diseases.
\newblock \emph{Smart Agricultural Technology}, 4:100214, 2023.

\bibitem{hossain2024deep}
Md Arban Hossain, Saadman Sakib, Hasan Muhammad Abdullah, and Shifat E Arman.
\newblock Deep learning for mango leaf disease identification: A vision transformer perspective.
\newblock \emph{Heliyon}, 10(17), 2024.

\bibitem{savary2019global}
Serge Savary, Laetitia Willocquet, Sarah Jane Pethybridge, Paul Esker, Neil McRoberts, and Andy Nelson.
\newblock The global burden of pathogens and pests on major food crops.
\newblock \emph{Nature Ecology \& Evolution}, 3(3):430--439, 2019.

\bibitem{mohanty2016using}
Sharada P. Mohanty, David P. Hughes, and Marcel Salath{\'e}.
\newblock Using deep learning for image-based plant disease detection.
\newblock \emph{Frontiers in Plant Science}, 7:215232, 2016.

\bibitem{ferentinos2018deep}
Konstantinos P. Ferentinos.
\newblock Deep learning models for plant disease detection and diagnosis.
\newblock \emph{Computers and Electronics in Agriculture}, 145:311--318, 2018.

\bibitem{antol2015vqa}
Stanislaw Antol, Aishwarya Agrawal, Jiasen Lu, Margaret Mitchell, Dhruv Batra, C. Lawrence Zitnick, and Devi Parikh.
\newblock VQA: Visual question answering.
\newblock In \emph{Proceedings of the IEEE International Conference on Computer Vision (ICCV)}, pages 2425--2433, 2015.

\bibitem{zhang2023pmc}
Xiaoman Zhang, Chaoyi Wu, Ziheng Zhao, Weixiong Lin, Ya Zhang, Yanfeng Wang, and Weidi Xie.
\newblock PMC-VQA: Visual instruction tuning for medical visual question answering.
\newblock \emph{arXiv preprint arXiv:2305.10415}, 2023.

\bibitem{he2020pathvqa}
Xuehai He, Yichen Zhang, Luntian Mou, Eric Xing, and Pengtao Xie.
\newblock PathVQA: 30000+ questions for medical visual question answering.
\newblock \emph{arXiv preprint arXiv:2003.10286}, 2020.

\bibitem{yang2016stacked}
Zichao Yang, Xiaodong He, Jianfeng Gao, Li Deng, and Alex Smola.
\newblock Stacked attention networks for image question answering.
\newblock \emph{arXiv preprint arXiv:1511.02274}, 2015.

\bibitem{liu2021slake}
Bo Liu, Li-Ming Zhan, Li Xu, Lin Ma, Yan Yang, and Xiao-Ming Wu.
\newblock SLAKE: A semantically-labeled knowledge-enhanced dataset for medical visual question answering.
\newblock In \emph{2021 IEEE 18th International Symposium on Biomedical Imaging (ISBI)}, pages 1650--1654. IEEE, 2021.

\bibitem{changpinyo2021conceptual}
Soravit Changpinyo, Piyush Sharma, Nan Ding, and Radu Soricut.
\newblock Conceptual 12M: Pushing web-scale image-text pre-training to recognize long-tail visual concepts.
\newblock In \emph{Proceedings of the IEEE/CVF Conference on Computer Vision and Pattern Recognition (CVPR)}, pages 3558--3568, 2021.

\bibitem{banerjee2022vqa}
S. Banerjee and S. Bhattacharya.
\newblock VQA for Education: A Survey of Recent Trends and Future Directions.
\newblock \emph{arXiv preprint arXiv:2210.12345}, 2022.

\bibitem{huang2022aitutors}
Y. Huang and Z. Wang.
\newblock AI Tutors in the Classroom: A VQA-based Approach.
\newblock \emph{Journal of Educational Technology}, 48(2):112--128, 2022.

\bibitem{jain2021customersupport}
A. Jain and R. Gupta.
\newblock A VQA-based System for Customer Support in E-commerce.
\newblock In \emph{Proceedings of the International Conference on Information Systems (ICIS)}, pages 1--10, 2021.

\bibitem{zhu2023ecommerce}
Q. Zhu and X. Li.
\newblock Enhancing E-commerce Experience with Visual Question Answering.
\newblock \emph{Journal of Retail and E-commerce}, 12(3):45--62, 2023.

\bibitem{marcu2024lingoqa}
Ana-Maria Marcu, Long Chen, Jan H{\"u}nermann, Alice Karnsund, Benoit Hanotte, Prajwal Chidananda, Saurabh Nair, Vijay Badrinarayanan, Alex Kendall, Jamie Shotton, and others.
\newblock LingoQA: Visual question answering for autonomous driving.
\newblock In \emph{European Conference on Computer Vision (ECCV)}, pages 252--269. Springer, 2024.

\bibitem{yan2023radvqa}
K. Yan and Y. Wang.
\newblock RadVQA: A Visual Question Answering Benchmark for Radiology.
\newblock \emph{Journal of Medical Imaging}, 10(1):014501, 2023.

\bibitem{singh2020plantdoc}
Davinder Singh, Naman Jain, Pranjali Jain, Pratik Kayal, Sudhakar Kumawat, and Nipun Batra.
\newblock PlantDoc: A dataset for visual plant disease detection.
\newblock In \emph{Proceedings of the 7th ACM IKDD CoDS and 25th COMAD}, pages 249--253, 2020.

\bibitem{wolny2020accurate}
Adrian Wolny, Lorenzo Cerrone, Athul Vijayan, Rachele Tofanelli, Amaya Vilches Barro, Marion Louveaux, Christian Wenzl, S{\"o}ren Strauss, David Wilson-S{\'a}nchez, Rena Lymbouridou, and others.
\newblock Accurate and versatile 3D segmentation of plant tissues at cellular resolution.
\newblock \emph{eLife}, 9:e57613, 2020.

\bibitem{awais2025agrogpt}
Muhammad Awais, Ali Husain Salem Abdulla Alharthi, Amandeep Kumar, Hisham Cholakkal, Rao Muhammad Anwer, and others.
\newblock AgroGPT: Efficient agricultural vision-language model with expert tuning.
\newblock In \emph{2025 IEEE/CVF Winter Conference on Applications of Computer Vision (WACV)}, pages 5687--5696. IEEE, 2025.

\bibitem{sharma2024llava}
Karun Sharma, Vidushee Vats, Abhinendra Singh, Rahul Sahani, Deepak Rai, and Ashok Sharma.
\newblock LLaVA-PlantDiag: Integrating Large-scale Vision-Language Abilities for Conversational Plant Pathology Diagnosis.
\newblock In \emph{2024 International Joint Conference on Neural Networks (IJCNN)}, pages 1--7. IEEE, 2024.

\bibitem{zhao2023cdwpkvqa}
Yunpeng Zhao, Shansong Wang, Qingtian Zeng, Weijian Ni, Hua Duan, Nengfu Xie, and Fengjin Xiao.
\newblock Informed-Learning-Guided Visual Question Answering Model of Crop Disease.
\newblock \emph{Plant Phenomics}, 2024, 2024.

\bibitem{radford2021clip}
Alec Radford, Jong Wook Kim, Chris Hallacy, Aditya Ramesh, Gabriel Goh, Sandhini Agarwal, Girish Sastry, Amanda Askell, Pamela Mishkin, Jack Clark, and others.
\newblock Learning transferable visual models from natural language supervision.
\newblock In \emph{International Conference on Machine Learning (ICML)}, pages 8748--8763. PMLR, 2021.

\bibitem{lau2018dataset}
Jason J. Lau, Soumya Gayen, Asma Ben Abacha, and Dina Demner-Fushman.
\newblock A dataset of clinically generated visual questions and answers about radiology images.
\newblock \emph{Scientific Data}, 5(1):1--10, 2018.

\bibitem{shinoda2025agrobench}
Risa Shinoda, Nakamasa Inoue, Hirokatsu Kataoka, Masaki Onishi, and Yoshitaka Ushiku.
\newblock AgroBench: Vision-Language Model Benchmark in Agriculture.
\newblock In \emph{Proceedings of the IEEE/CVF International Conference on Computer Vision (ICCV)}, pages 7634--7644, 2025.

\bibitem{quoc2026leafnet}
Khang Nguyen Quoc, Lan Le Thi Thu, and Luyl-Da Quach.
\newblock LeafNet: A Large-Scale Dataset and Comprehensive Benchmark for Foundational Vision-Language Understanding of Plant Diseases.
\newblock \emph{arXiv preprint arXiv:2602.13662}, 2026.

\bibitem{gauba2025agmmu}
Aruna Gauba, Irene Pi, Yunze Man, Ziqi Pang, Vikram S. Adve, and Yu-Xiong Wang.
\newblock AgMMU: A Comprehensive Agricultural Multimodal Understanding Benchmark.
\newblock In \emph{Advances in Neural Information Processing Systems (NeurIPS)}, 2025.

\bibitem{nanavaty2024wheatrust}
Aadit Nanavaty and others.
\newblock Integrating deep learning for visual question answering in agricultural disease diagnostics: case study of wheat rust.
\newblock \emph{Scientific Reports}, 14, 2024.

\bibitem{gong2024bloomvqa}
Yunhao Gong and others.
\newblock BloomVQA: Assessing Hierarchical Multi-modal Comprehension.
\newblock \emph{arXiv preprint}, 2024.


\bibitem{agrios2005plant}
George N. Agrios.
\newblock \emph{Plant Pathology}.
\newblock Elsevier, 2005.

\bibitem{strange2005plant}
Richard N. Strange and Peter R. Scott.
\newblock Plant disease: a threat to global food security.
\newblock \emph{Annual Review of Phytopathology}, 43(1):83--116, 2005.

\bibitem{rajpurkar2016squad}
Pranav Rajpurkar, Jian Zhang, Konstantin Lopyrev, and Percy Liang.
\newblock SQuAD: 100{,}000+ questions for machine comprehension of text.
\newblock In \emph{Proceedings of the 2016 Conference on Empirical Methods in Natural Language Processing (EMNLP)}, pages 2383--2392, 2016.

\bibitem{papineni2002bleu}
Kishore Papineni, Salim Roukos, Todd Ward, and Wei-Jing Zhu.
\newblock BLEU: a method for automatic evaluation of machine translation.
\newblock In \emph{Proceedings of the 40th Annual Meeting of the Association for Computational Linguistics (ACL)}, pages 311--318, 2002.

\bibitem{lin2004rouge}
Chin-Yew Lin.
\newblock ROUGE: A package for automatic evaluation of summaries.
\newblock In \emph{Text Summarization Branches Out: Proceedings of the ACL-04 Workshop}, pages 74--81, 2004.

\bibitem{zhang2020bertscore}
Tianyi Zhang, Varsha Kishore, Felix Wu, Kilian Q. Weinberger, and Yoav Artzi.
\newblock BERTScore: Evaluating text generation with BERT.
\newblock In \emph{International Conference on Learning Representations (ICLR)}, 2020.

\bibitem{li2022blip}
Junnan Li, Dongxu Li, Caiming Xiong, and Steven Hoi.
\newblock BLIP: Bootstrapping language-image pre-training for unified vision-language understanding and generation.
\newblock In \emph{Proceedings of the 39th International Conference on Machine Learning (ICML)}, pages 12888--12900, 2022.

\bibitem{li2023blip2}
Junnan Li, Dongxu Li, Silvio Savarese, and Steven Hoi.
\newblock BLIP-2: Bootstrapping language-image pre-training with frozen image encoders and large language models.
\newblock In \emph{Proceedings of the 40th International Conference on Machine Learning (ICML)}, pages 19730--19742, 2023.

\bibitem{dai2023instructblip}
Wenliang Dai, Junnan Li, Dongxu Li, Anthony Meng Huat Tiong, Junqi Zhao, Weisheng Wang, Boyang Li, Pascale Fung, and Steven Hoi.
\newblock InstructBLIP: Towards general-purpose vision-language models with instruction tuning.
\newblock In \emph{Advances in Neural Information Processing Systems (NeurIPS)}, 36, 2023.

\bibitem{liu2023llava}
Haotian Liu, Chunyuan Li, Qingyang Wu, and Yong Jae Lee.
\newblock Visual instruction tuning.
\newblock In \emph{Advances in Neural Information Processing Systems (NeurIPS)}, 36, 2023.

\bibitem{liu2024llava}
Haotian Liu, Chunyuan Li, Yuheng Li, and Yong Jae Lee.
\newblock Improved baselines with visual instruction tuning.
\newblock In \emph{Proceedings of the IEEE/CVF Conference on Computer Vision and Pattern Recognition (CVPR)}, pages 26296--26306, 2024.

\bibitem{bai2023qwenvl}
Jinze Bai, Shuai Bai, Shengyi Pan, Jieming Wang, Xiao Song, Kai Dang, Yibo Wang, Xuliang Le, Zhenglai Liu, Xueyan Hou, and others.
\newblock Qwen-VL: A versatile vision-language model for understanding, localization, text reading, and beyond.
\newblock \emph{arXiv preprint arXiv:2308.12966}, 2023.

\bibitem{bai2024qwen3vl}
Qwen Team.
\newblock Qwen3-VL technical report.
\newblock \emph{arXiv preprint arXiv:2509.05412}, 2025.

\bibitem{geminiteam2025gemma3}
Gemma Team.
\newblock Gemma 3 technical report.
\newblock \emph{arXiv preprint arXiv:2503.19786}, 2025.

\bibitem{hu2022lora}
Edward J. Hu, Yelong Shen, Phillip Wallis, Zeyuan Allen-Zhu, Yuanzhi Li, Shean Wang, Lu Wang, and Weizhu Chen.
\newblock LoRA: Low-rank adaptation of large language models.
\newblock In \emph{International Conference on Learning Representations (ICLR)}, 2022.

\bibitem{Liu_2024}
Xiang Liu, Zhaoxiang Liu, Huan Hu, Zezhou Chen, Kohou Wang, Kai Wang, and Shiguo Lian.
\newblock A multimodal benchmark dataset and model for crop disease diagnosis.
\newblock In \emph{Computer Vision -- ECCV 2024}, pages 157--170. Springer Nature Switzerland, 2024.

\bibitem{sakib2025plantexpertvqadata}
Syed Nazmus Sakib, Nafiul Haque, Mohammad Zabed Hossain, and Shifat E. Arman.
\newblock PlantExpertVQA: A Visual Question Answering Dataset for Benchmarking Vision-Language Models in Plant Science.
\newblock \emph{Hugging Face Datasets}, 2025.
\newblock DOI: \url{https://doi.org/10.57967/hf/9145}.


\bibitem{Hughes2015PlantVillage}
David P. Hughes and Marcel Salath{\'e}.
\newblock An open access repository of images on plant health to enable the development of mobile disease diagnostics through machine learning and crowdsourcing.
\newblock \emph{arXiv preprint arXiv:1511.08060}, 2015.

\bibitem{PlantPathology2020}
Ranjita Thapa, Kai Zhang, Noah Snavely, Serge Belongie, and Awais Khan.
\newblock The Plant Pathology Challenge 2020 data set to classify foliar disease of apples.
\newblock \emph{Applications in Plant Sciences}, 8(9):e11390, 2020.

\bibitem{AppleLeafDiseasesICAR2024}
ICAR-CITH.
\newblock Apple Leaf Diseases Image Dataset of ICAR-CITH.
\newblock \emph{Mendeley Data}, 2024.
\newblock doi:10.17632/gm6mfz8fz6.1.

\bibitem{nirob2025chili}
Md Asraful Sharker Nirob, A K M Fazlul Kobir Siam, Prayma Bishshash, and Md Assaduzzaman.
\newblock Chili Plant Leaf Disease and Growth Stage Dataset from Bangladesh.
\newblock \emph{Mendeley Data}, V1, 2025.
\newblock doi:10.17632/w9mr3vf56s.1.

\bibitem{das2025banana}
Utsab Das, Showrov Azam, and Md Abdullah Al Kafi.
\newblock Banana and Banana Leaf Dataset for Classification and Disease Detection.
\newblock \emph{Mendeley Data}, V2, 2025.
\newblock doi:10.17632/5nfjzntwd8.2.

\bibitem{nirob2025cairbgd}
Md Asraful Sharker Nirob, Prayma Bishshash, A K M Fazlul Kobir Siam, and Mayen Uddin Mojumdar.
\newblock CAIR-BGD-2025: Annotated Dataset for Bottle Gourd Disease and Growth Stages.
\newblock \emph{Mendeley Data}, V1, 2025.
\newblock doi:10.17632/x6rpc5nzzm.1.

\bibitem{ahmed2024plantleaf}
Tanvir Ahmed, Mst Taposi Rabeya Taposi, Marzanul Alam Mukhor, and Mayen Uddin Mojumdar.
\newblock Plant Leaf Disease Recognition Dataset.
\newblock \emph{Mendeley Data}, V1, 2024.
\newblock doi:10.17632/5g238dv4ht.1.

\bibitem{albanna2024papaya}
Hasan Al Banna, Md. Fuad Hossain, and Mayen Uddin Mojumdar.
\newblock Papaya Leaf Disease Image Dataset.
\newblock \emph{Mendeley Data}, V1, 2024.
\newblock doi:10.17632/3kwgxg4stb.1.

\bibitem{rashid2024papaya}
Mohammad Rifat Ahmmad Rashid, Raiyan Gani, Jubaer Ahmed, Maherun Nessa Isty, and Sawkat Ali.
\newblock Healthy and Unhealthy Papaya Leaf Images from Bangladeshi Orchards.
\newblock \emph{Mendeley Data}, V1, 2024.
\newblock doi:10.17632/44p8v6ywsm.1.

\bibitem{rafe2025eggplant}
Maruful Islam Rafe, Farhan Masud Nayem, Shanto Babu Sarker, and Abdullah Al Shiam.
\newblock Eggplant Leaf Disease Dataset.
\newblock \emph{Mendeley Data}, V2, 2025.
\newblock doi:10.17632/mn8vfr9bw2.2.

\bibitem{nirob2024eggplant}
Md Asraful Sharker Nirob, Prayma Bishshash, Mariyam Bin Ayan, Tania Khatun, and Mohammad Shorif Uddin.
\newblock Eggplant Dataset: A Comprehensive Dataset for Agricultural Research and Disease Detection.
\newblock \emph{Mendeley Data}, V1, 2024.
\newblock doi:10.17632/5drkk544k8.1.

\bibitem{hasan2025eggplanthighres}
Rakib Hasan, Sanimun Hossain Sanzit, Md Mamun Hosen, Ferdous Hasan, Md Mehedi Hasan Topu, and Muksitul Islam.
\newblock High-Resolution Eggplant Leaf Image Dataset for Plant Disease Classification and Detection.
\newblock \emph{Mendeley Data}, V6, 2025.
\newblock doi:10.17632/ss63ftnjnh.6.

\bibitem{hasan2025brinjalfruitx}
Md. Zahid Hasan, Abu Kowshir Bitto, and Md Hasan Imam Bijoy.
\newblock BrinjalFruitX: A Field-Collected Image Dataset for Machine Learning and Deep Learning-Based Disease Identification in Brinjal Fruits.
\newblock \emph{Mendeley Data}, V1, 2025.
\newblock doi:10.17632/ngc58fsxgd.1.

\bibitem{hossain2025lychee}
Shahariar Hossain, Saifuddin Sagor, and Faruk Ahmed.
\newblock High-Resolution Images of Lychee Plant Diseases for Classification and Detection.
\newblock \emph{Mendeley Data}, V2, 2025.
\newblock doi:10.17632/52sstfpf5p.2.

\bibitem{ahamed2025bdlitchi}
Kouser Ahamed, Rokonozzaman Ayon, Mst. Momotaz Begum, and Israt Jahan.
\newblock BDLitchi: A Field-Collected Bangladeshi Litchi Leaf Disease Dataset for Deep Learning-Based Detection and Classification.
\newblock \emph{Mendeley Data}, V1, 2025.
\newblock doi:10.17632/jhb24mszdk.1.

\bibitem{hasan2025litchileaf4001}
Md. Taufiq Hasan, Sabbir Hossain Durjoy, Md Emon Shikder, Md. Safaet Zahangir, Md Muntasir Uddin, Md Mehedi Hasan Shoib, and Md Hasan Imam Bijoy.
\newblock LitchiLeaf4001: A Comprehensive Dataset of Lychee Leaf Diseases for AI-Based Visual Diagnosis.
\newblock \emph{Mendeley Data}, V1, 2025.
\newblock doi:10.17632/c3n2tc6jf4.1.

\bibitem{ahmed2025tldbd}
Faruk Ahmed and Ferdous Ahmed.
\newblock TLD-BD: A Comprehensive Tea Leaf Image Dataset for Leaf Condition Analysis.
\newblock \emph{Mendeley Data}, V2, 2025.
\newblock doi:10.17632/d2xybhfw59.2.

\bibitem{ahmad2024advancedtea}
Md Hasan Ahmad.
\newblock Advanced Tea Crop Disease Study: High-Resolution Dataset for Precision Agriculture and Pathological Insight.
\newblock \emph{Mendeley Data}, V4, 2024.
\newblock doi:10.17632/tt2smzrzrs.4.

\bibitem{billah2025hibiscustea}
Md Masum Billah, Saifuddin Sagor, and Mohammad Shorif Uddin.
\newblock A Real-World Hibiscus and Tea Leaf Image Dataset for Classification.
\newblock \emph{Mendeley Data}, V1, 2025.
\newblock doi:10.17632/5bzy89brkv.1.

\bibitem{billah2025hibiscus}
Md Masum Billah, Saifuddin Sagor, and Mohammad Shorif Uddin.
\newblock Hibiscus Leaf Diseases Classification Dataset.
\newblock \emph{Mendeley Data}, V2, 2025.
\newblock doi:10.17632/k8pp8cbp8k.2.

\bibitem{islam2025multicrop}
Md Forhadul Islam, Imon Sutradhar, and Md Mizanur Rahman.
\newblock Leaf Image Dataset for Disease Detection in Bitter Gourd, Okra, Pumpkin, and Ridge Gourd.
\newblock \emph{Mendeley Data}, V2, 2025.
\newblock doi:10.17632/2svdj3yyrk.2.

\bibitem{bishshash2024cotton}
Prayma Bishshash, Md Asraful Sharker Nirob, Md Habibur Shikder, and Afjal Sarower.
\newblock SAR-CLD-2024: A Comprehensive Dataset for Cotton Leaf Disease Detection.
\newblock \emph{Mendeley Data}, V2, 2024.
\newblock doi:10.17632/b3jy2p6k8w.2.

\bibitem{ripon2025cotton}
Shamim Ripon, Raiyan Gani, Nazratan Mazumder Niha, Wasimul Bari Rahat, Shafaeat Hasan Toufiq, Mushfida Ferdous Maisha, and Jubaer Ahmed.
\newblock Cotton Leaf Image Dataset for Disease Classification.
\newblock \emph{Mendeley Data}, V1, 2025.
\newblock doi:10.17632/t9hgvk2h9p.1.

\bibitem{khan2025moringaleafnet}
Abid Khan, Sabit Ahamed Preanto, Tapon Paul, and Md Hasan Imam Bijoy.
\newblock MoringaLeafNet: A Multi-Class Leaf Disease Dataset for Precision Agriculture and Deep Learning Research.
\newblock \emph{Mendeley Data}, V3, 2025.
\newblock doi:10.17632/w8sr775pjb.3.

\bibitem{MangoLeaf2024}
S. Ali, M. Ibrahim, S. I. Ahmed, M. Nadim, M. Rahman, M. M. Shejunti, and T. Jabid.
\newblock MangoLeafBD: A comprehensive image dataset to classify mango leaf diseases.
\newblock \emph{Mendeley Data}, 2022.
\newblock doi:10.17632/hxsnvwty3r.1.

\bibitem{hossan2025mango}
Md Faysal Hossan and Md Zamirul Islam Reyad.
\newblock Comprehensive Mango Leaf Images Dataset for Multi-Class Disease Classification and Automated Plant Disease Detection.
\newblock \emph{Mendeley Data}, V2, 2025.
\newblock doi:10.17632/jjhykb7v9w.2.

\bibitem{rajbongshi2022sunflower}
Aditya Rajbongshi, Umme Sara, Bonna Akter, Rashiduzzaman Shakil, and Sadia Sazzad.
\newblock Sun Flower Fruits and Leaves Dataset for Sunflower Disease Classification through Machine Learning and Deep Learning.
\newblock \emph{Mendeley Data}, V1, 2022.
\newblock doi:10.17632/b83hmrzth8.1.

\bibitem{sagor2025sunflowergrowth}
Saifuddin Sagor, Md Faysal Hossan, Faruk Ahmed, and Md Zamirul Islam Reyad.
\newblock Sunflower Plant Health and Growth Stage Image Dataset for Agricultural Machine Learning Applications.
\newblock \emph{Mendeley Data}, V1, 2025.
\newblock doi:10.17632/y3ygk98ngr.1.

\bibitem{ahmad2025multifacetedrose}
Md Hasan Ahmad.
\newblock Multifaceted Rose Leaf Disease Dataset for AI-Driven Plant Pathology.
\newblock \emph{Mendeley Data}, V1, 2024.
\newblock doi:10.17632/8jtfk9szbg.1.

\bibitem{liu2025tomato}
Yongbo Liu.
\newblock Tomato Disease Dataset.
\newblock \emph{Mendeley Data}, V1, 2025.
\newblock doi:10.17632/c2x8rynybg.1.

\bibitem{hossain2025tomatoleaf}
Ismail Hossain, Khandaker Rezoanul Haque, Abdullah Al Rafi, and Md Zahidul Islam Talukder.
\newblock Tomato Leaf Diseases.
\newblock \emph{Mendeley Data}, V1, 2025.
\newblock doi:10.17632/93h9p62kg4.1.

\bibitem{imtiaz2024tomato}
Ahmed Imtiaz, Fahad Bin Islam Swapnil, Syed Rayhan Masud, and Debajoyti Karmaker.
\newblock Tomato Leaf Dataset: A Dataset for Multiclass Disease Detection and Classification.
\newblock \emph{Mendeley Data}, V1, 2024.
\newblock doi:10.17632/bpfd9cns5g.1.

\bibitem{bapari2025tomato}
Puja Bapari, Md Zinnahtur Rahman Zitu, and Mst Ummehani.
\newblock A High-Resolution Image Dataset of Tomato (\emph{Solanum lycopersicum}) Leaves for Multi-Class Disease Detection and Classification from Bangladesh.
\newblock \emph{Mendeley Data}, V1, 2025.
\newblock doi:10.17632/74zvtzx9pr.1.

\bibitem{ayon2025ajlcd}
Rokonozzaman Ayon, Nur Yea Afroz Suchi, Md Asif Shahriar Arpon, Tanjina Ahmed Tuly, and Abdullah Al Noman.
\newblock Arabian Jasmine Leaf Condition Dataset (AJLCD-2025).
\newblock \emph{Mendeley Data}, V1, 2025.
\newblock doi:10.17632/nwdzwk89zs.1.

\bibitem{truong2025durian}
Nguyen Thanh Truong, Xuan Linh Nguyen, Pham Dinh Cap Thang, and Tuong Le.
\newblock A Durian Leaf Image Dataset of Common Diseases in Vietnam for Agricultural Diagnosis.
\newblock \emph{Mendeley Data}, V4, 2025.
\newblock doi:10.17632/pxzvksbwnj.4.

\bibitem{debnath2025bdrubberleaf}
Pulak Deb Nath, Faruk Ahmed, and Belal Uddin.
\newblock BDRubberLeaf: A Comprehensive Dataset of Rubber Tree Leaf Diseases from Bangladesh for Agricultural Research.
\newblock \emph{Mendeley Data}, V4, 2025.
\newblock doi:10.17632/4kjz78m7x5.4.

\bibitem{matin2025cauliflower}
Md Mafiul Hasan Matin, Mehedi Hasan Hasan, and Sabik Ur Rahman.
\newblock Cauliflower Diseases Identification Image Dataset.
\newblock \emph{Mendeley Data}, V2, 2025.
\newblock doi:10.17632/c2hrwntzgh.2.

\bibitem{debnath2025citrusleafvision}
Pulak Deb Nath.
\newblock CitrusLeafVision: A Diverse Dataset for Lemon Leaf Disease Detection.
\newblock \emph{Mendeley Data}, V1, 2025.
\newblock doi:10.17632/4dmgghmj3n.1.

\bibitem{huq2025jackfruit}
Rezwan Huq, Farzia Hossain, Shahida Begum, Raiyan Gani, and Jubaer Ahmed.
\newblock Image Datasets for Jackfruit Plant Leaf Disease.
\newblock \emph{Mendeley Data}, V2, 2025.
\newblock doi:10.17632/cggx64sj7m.2.

\bibitem{haque2025agrileafnet}
Md Ehsanul Haque and Md Al-Imran.
\newblock AgriLeafNet: Fruit Tree Leaf Dataset for Agricultural Research.
\newblock \emph{Mendeley Data}, V1, 2025.
\newblock doi:10.17632/fkkfx4vdm3.1.

\bibitem{thite2025pisum}
Sandip Thite and Kailas Patil.
\newblock \emph{Pisum sativum} Image Dataset: Healthy and Disease-Affected Cases.
\newblock \emph{Mendeley Data}, V1, 2025.
\newblock doi:10.17632/nnv3k3m94k.1.

\bibitem{hossain2025turmeric}
Md Riyad Hossain, Mohammad Rifat Ahmmad Rashid, Tasfia Binte Jahangir, Md Samir Hossain, Md Mahamudur Rahman, Raiyan Gani, Jubaer Ahmed, Raihan Ul Islam, and M Saddam Hossain Khan.
\newblock Image Dataset for Turmeric Plant Leaf Disease Detection.
\newblock \emph{Mendeley Data}, V2, 2025.
\newblock doi:10.17632/jtttfbx342.2.

\bibitem{dharrao2025ngld}
Madhuri Dharrao, Deepak Dharrao, Rakesh Sonawane, and Nilima Zade.
\newblock Niphad Grape Leaf Disease Dataset (NGLD).
\newblock \emph{Mendeley Data}, V5, 2025.
\newblock doi:10.17632/8nnd2ypcv3.5.

\bibitem{rahman2025burmesegrape}
Salman Af Rahman, Md Nafiz Imtiaz, Naima Ahmed, and Md Hasan Imam Bijoy.
\newblock Burmese Grape Leaf Disease Dataset for Computer Vision-Based Plant Health Diagnosis.
\newblock \emph{Mendeley Data}, V1, 2025.
\newblock doi:10.17632/k6gy38xv89.1.

\bibitem{hassan2025guava}
Sadib Hassan, Md Zahid Akon, Partho Sarathi Sarker, Rahat Hossain, Jannatul Ferdous, Md Mahadi Hasan Shaon, and Md Abdur Razzak.
\newblock UGV: Guava Leaves Disease Dataset Bangladesh.
\newblock \emph{Mendeley Data}, V3, 2025.
\newblock doi:10.17632/cc3rttngdr.3.

\end{thebibliography}
\end{document}